\documentclass[runningheads]{llncs}

\usepackage{eccv}

\usepackage{eccvabbrv}

\usepackage{graphicx}
\usepackage{booktabs}

\usepackage[accsupp]{axessibility}  

\usepackage{amsmath}
\usepackage{makecell}

\usepackage{hyperref}

\usepackage{orcidlink}

\begin{document}

\title{Topology-Preserving Downsampling of Binary Images}

\titlerunning{Topology-Preserving Downsampling of Binary Images}

\author{Chia-Chia Chen\inst{1, 2}\orcidlink{0009-0005-5552-1625} \and
Chi-Han Peng\inst{1,3}\orcidlink{0000-0002-6823-8029}}

\authorrunning{CC Chen, CH Peng}

\institute{College of AI, National Yang Ming Chiao Tung University, Taiwan \and
\email{s1073736@gmail.com} \and
\email{pengchihan@nycu.edu.tw}}

\maketitle

\begin{abstract}
  
  We present a novel discrete optimization-based approach to generate downsampled versions of binary images that are guaranteed to have the same topology as the original, measured by the zeroth and first Betti numbers of the black regions, while having good similarity to the original image as measured by IoU and Dice scores. To our best knowledge, all existing binary image downsampling methods don't have such topology-preserving guarantees. We also implemented a baseline morphological operation (dilation)-based approach that always generates topologically correct results. However, we found the similarity scores to be much worse. We demonstrate several applications of our approach. First, generating smaller versions of medical image segmentation masks for easier human inspection. Second, improving the efficiency of binary image operations, including persistent homology computation and shortest path computation, by substituting the original images with smaller ones. In particular, the latter is a novel application that is made feasible only by the full topology-preservation guarantee of our method.
\end{abstract}

\section{Introduction}
\label{sec:intro}

Binary images, i.e., images consisting of only "foreground" (e.g., black) and "background" (e.g., white) pixels, are widely used in computer visions and computer graphics. Example usages include data structures for segmentation masks, image inputs to document analysis and industry machine vision systems\cite{shaprio01}, and the encoding of 2D geographical maps in games and animations. A classic book on machine vision~\cite{jain1995machine} pointed to several advantages of binary images. First, designers noted that binary images are easier for humans to recognize key features such as silhouettes. Second, binary images led to significantly smaller memory and processing requirements of computer vision algorithms than their grey-level or color image counterparts. In particular, the binary nature of pixel values enables efficient run-length encoding of the data storage and the applicability of binary logical operations instead of integer arithmetic operations.

A key characteristic of binary images compared to grey-level or color images is their natural correspondence to 2D graphs. To do so, we first convert a binary image to a regular quad mesh by taking every pixel as a quad face. We then remove all quads corresponding to white pixels and the resulting dangling vertices and edges (i.e., vertices and edges without adjacent faces). We now have a graph consisting of black-pixel quads only. This graph has zero or more connected components and zero or more holes inside each components. We denote the number of connected components and the total number of holes, namely the zeroth and first Betti numbers, as the {\em topology} of the binary image.

The topology of binary images plays a key role in their applications. There exists a large body of research concerning the efficient computation of topological operations on binary images such as border following, region adjacency graph, medial axis / skeletonization calculation, morphological operations (such as thinning and dilation), and distance transforms~\cite{yokoi1975analysis, suzuki1985topological,shaprio01,jain1995machine,abu2013skeletonization,boudaoud2015new,wang2021spline,strutz2021distance}. For neural image and volume segmentation methods, it has been shown that incorporating prior knowledge about the topology (in terms of Betti numbers) of the segmented objects, usually in forms of new loss functions, can improve the segmentation results' topological accuracy and in some cases even per-pixel accuracy~\cite{clough2020topological,hu2019topology,hu2022structure}. In all these applications, having topologically correct versions of a binary image at different resolutions can be useful.


\begin{figure}[t]
  \centering
  \includegraphics[width=1\linewidth]{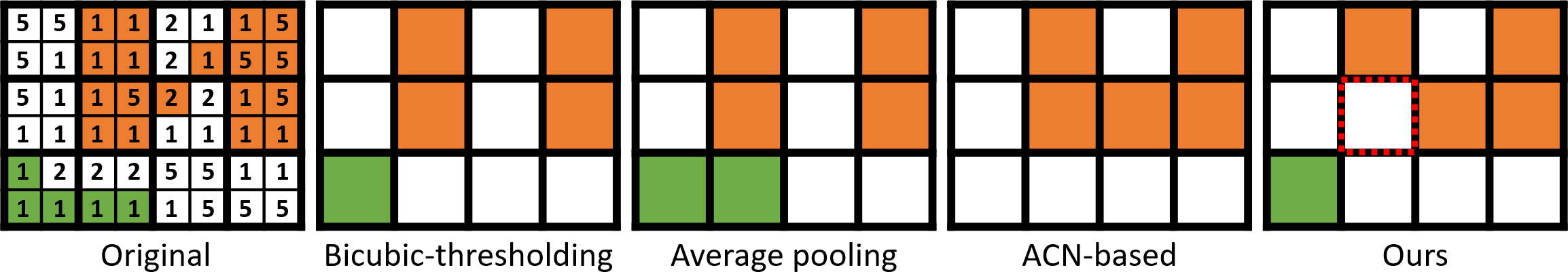}
  \caption{Downsampling a binary image of two connected components by a factor of 2 (i.e., turning each 2x2 block of pixels into a downsampled pixel). Bicubic filtering-and-thresholding (by OpenCV), average pooling (when tie choose foreground color), and the adaptive crossing number (ACN)-based method~\cite{decenciere2007adaptive}, all generate results with different topology. We show the ACNs in each original pixel. Our method finds a topologically correct result. A non-trivial pixel-color decision is marked in red.}
  \label{fig:difficult_cases}
\end{figure}

\begin{figure}[t]
  \centering
  \includegraphics[width=0.97\linewidth]{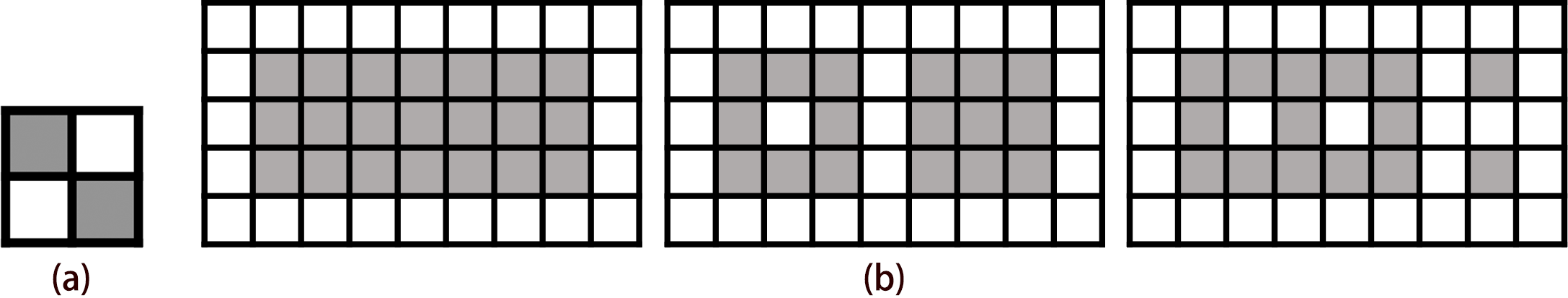}
  \caption{(a) A binary image without any topologically correct downsampling solutions. (b) Three black regions with different Betti numbers (from left to right: (1,0), (2,1), (3,2)). However, their Euler characteristics all equal to 1. This can be verified by their numbers of vertices, edges, and faces: (32,52,21), (32,48,17), and (32,46,15). This is expected because Euler characteristic equals the number of connected components minus the number of holes for 2D graphs (\cite{jain1995machine}).}
  \label{fig:impossible_case_and_Euler}
\end{figure}

However, preserving the topology when downsampling a binary image is a non-trivial and sometimes impossible task. To our best knowledge, all existing binary image downsampling methods, including bilinear or bicubic filtering-and-thresholding, various pooling-based approaches, and an adaptive crossing number-based method proposed in 2007~\cite{decenciere2007adaptive}, can cause topological changes. We show examples in Figure~\ref{fig:difficult_cases}. There, we can see that the color of a downsampled pixel can not be decided by simply checking the ratios of black or white pixels in its corresponding block of original pixels - {\em sometimes a downsampled pixel has to be white even if its corresponding block of pixels are all black}. In Figure~\ref{fig:impossible_case_and_Euler} (a), we show a binary image without any topological correct downsample solutions. Note that our approach finds a non-trivial and correct solution to Figure~\ref{fig:difficult_cases} and definitely declares Figure~\ref{fig:impossible_case_and_Euler} (a) to be an infeasible problem. 

One can instead design a downsampling algorithm that guarantees to output topologically correct results by leveraging morphological operations. Although we did not find such algorithms published in research venues, we envision a baseline approach as follows. For each connected component in the original binary image, we create a corresponding trivial shape with the same topology in the downsampled image. We then iteratively improve the shapes of each component by dilation operations. We found that such a greedy approach can not produce results that have high similarity to the original binary image.

In short, the downsampling task has two main goals - the result binary image shall have the same topology and also high similarity to the original. Our insight is to formulate the task as an {\em discrete optimization} problem - in which the black-or-white decisions of each downsampled pixels are encoded as Boolean variables, the topology preservation is a hard constraint, and the similarity to the original binary image is the objective function to maximize. In this way, any computed solution to the optimization problem is akin to a topologically correct downsampled binary image while having good similarity to the original. Another benefit of such a optimization-based approach is that impossible tasks are identified as infeasible problems.

The bottleneck of the optimization problem design is the formulation of the topology-preserving hard constraint. As has been shown in Figure~\ref{fig:difficult_cases}, topology-preservation is a global phenomena where the decision of one downsampled pixel can be influenced by regions far away. Therefore, any local approaches (i.e., making the decision based on the values of nearby pixels) would not work. We also found that constraining the Euler characteristics of each components does not work because two graphs with different Betti numbers can have the same Euler characteristic (shown in Figure~\ref{fig:impossible_case_and_Euler} (b)). Inspired by the topological "no-island" constraint for network design problems in~\cite{10.1145/2897824.2925935}, we propose a novel formulation to ensure that {\em the topology of each component's boundary to remain a circle}, which subsequently ensures that the topology of each component to remain the same. 

In summary, our contributions are as follows:

\begin{itemize}
  \item We propose a novel discrete optimization-based approach to tackle the previously unsolved topology-preserving binary image downsampling problem. Our method outputs results that are not only topologically correct, but also have good similarity scores that are competitive with conventional approaches without topology-preservation guarantees.
  \item We have designed our problem formulation such that it can be solved efficiently. In fact, our method can be used to speed up binary image operations such as persistent homology computation.
  \item We show that the main advantage of our method, namely full topology-preservation, enables novel applications of binary image downsampling that traditional downsampling methods could not support.
\end{itemize}

\section{Related Work}
\label{sec:related}

\subsection{Existing Binary Image Downsampling Methods}

We base our discussions of existing common binary image downsampling methods on the implementations in OpenCV~\cite{opencv_library}. To find the color of a downsampled pixel at coordinate $(X,Y)$, the \textbf{Nearest Neighbor} approach simply converts $(X,Y)$ back to the original resolution, $(A*X, B*Y)$, $A$ and $B$ are the downsampling factors in width and height, and use the pixel color at $(A*X, B*Y)$. \textbf{Average Pooling} (in OpenCV it is called "\textbf{Area}") assigns black to a downsampled pixel if the ratio of black pixels in its $A$-by-$B$ block equal or greater than half, and assigns white otherwise. Similarly, \textbf{Max or Min Pooling} choose a black big-pixel if any or all of the corresponding pixels are black, respectively. \textbf{Bilinear or Bicubic Interpolation-And-Thresholding} does a grey-image downsampling using the standard bilinear or bicubic interpolation first, and then convert the grey-level result to a binary image using thresholding at half grey.


In~\cite{Ngo2014} and~\cite{6690186}, the necessary and sufficient conditions for a binary image to retain topology after arbitrary rigid transformations (translation and rotation, but no scaling) are analyzed. In~\cite{passat:hal-03630330}, an algorithm to do arbitrary topology-invariant affine transformations (including scaling) for binary images is proposed. However, the algorithm often could not reach a feasible solution (if one exists) for downsampling tasks. A detailed discussion is in the Supplementary Materials.

The adaptive crossing number (ACN)-based method~\cite{decenciere2007adaptive} also aimed for binary image downsampling with topology-preservation as a goal. However, the method does not guarantee topologically correct results. The method only performs downsampling by a factor of 2. In short, ACN are integer numbers assigned to each pixels in the original image such that higher values indicate pixels that are more likely to alter the topology when not chosen, and vice versa. For each downsampled pixel, the color of the pixel with the highest ACN is chosen. In tie, the pixel in the first scanning order (the upper-left corner) is chosen.

\subsection{Leveraging Topology for Neural Image  Segmentation Methods}

There exist many methods that leverage the topological properties of segmentation masks to improve the results of neural segmentation methods. They mostly do so by introducing novel loss functions that are more sensitive to the topological differences between prediction results and the ground truth. In~\cite{mosinska2018beyond}, the authors found that the feature maps of a pretrained VGG19 neural network~\cite{Simonyan15} tend to be more topologically correct and designed a topological loss function using the outputs of these feature maps. In~\cite{hu2019topology}, a more direct and effective topological loss function is introduced that measures the differences in terms of the persistent homology (PH) of segmentation masks. Note that this method encodes only the first Betti number (numbers of connected components) of 2D segmentation masks. In~\cite{clough2020topological}, a more general PH-based cost function design is introduced that can encode arbitrary lengths of Betti numbers of segmented objects (e.g., numbers of cavities in volumetric data). In~\cite{hu2022structure}, a non PH-based cost function design, based on homotopy warping, is introduced that showed improved accuracy and computational cost. Other non PH-based topological cost function designs include one based on morphological skeleta~\cite{shit2021cldice} and one based on Euler characteristics~\cite{li2023robust}. Our method can benefit these methods by providing easier data visualization. Also, we envision that being able to quickly generate segmentation masks at reduced resolutions but still with the same topology (which was not possible prior to our method) can lead to novel algorithm designs.

\noindent\textbf{Persistent Homology (PH).} In short, PH encodes the "birth" and "death" times of each components during a consecutive session of dilation from each component being a single point until all components are so inflated to merge into a single component without holes. Unlike Betti numbers which are discrete, PH are continuous-valued and differentiable encoding. Therefore, they can be easily modeled as cost functions of neural networks. However, the main downside is that PH computation is expensive (\cite{hu2022structure,li2023robust}). In Section~\ref{sec:result_speed}, we demonstrate that our method can be used to significantly reduce the cost of PH computation with a small impact to its accuracy.

\section{Our Method}
\label{sec:method}

We begin with definitions. A binary image consists of pixels of exactly two possible colors - foreground ("black" in short) and background ("white" in short). Note that in our figures we may use different colors to denote pixels of different connected components (e.g., Figure~\ref{fig:difficult_cases}). We downsample a binary image of width $W$ and height $H$ by a factor of $A$ in width and $B$ in height. We only consider the case of $A$ and $B$ being positive integers and $W$ is dividable by $A$ and $H$ is dividable by $B$. In other words, we practically convert each $A$-by-$B$ block of pixels of possibly different colors into one downsampled "big"-pixel of a single color. The resolution of the downsampled binary image is $W/A$-by-$H/B$.

A $W$-by-$H$ binary image can be turned into a regular grid-like quad mesh of $H$ rows and $W$ columns of quads. A quad is black or white if it  corresponds to a black or white pixel, respectively. If we remove all the white quads and the resulting "dangling" vertices and edges (i.e., those without any adjacent faces), we are left with a quad mesh of black quads only. Note that we consider an edge is adjacent to a face if and only if that face has the edge as one of its sides. This {\em black-quad mesh} of the binary image may have zero or more connected components. Note that we consider two black quads sharing just a single vertex (i.e., in a diagonal position) as neighbors. In each of the connected component, there may be zero or more holes. We denote the numbers of connected components and the total numbers of holes, namely the zeroth and first Betti numbers of the black-quad mesh, as the {\em topology} of the binary image.

\begin{figure}[t]
  \centering
  \includegraphics[width=1\linewidth]{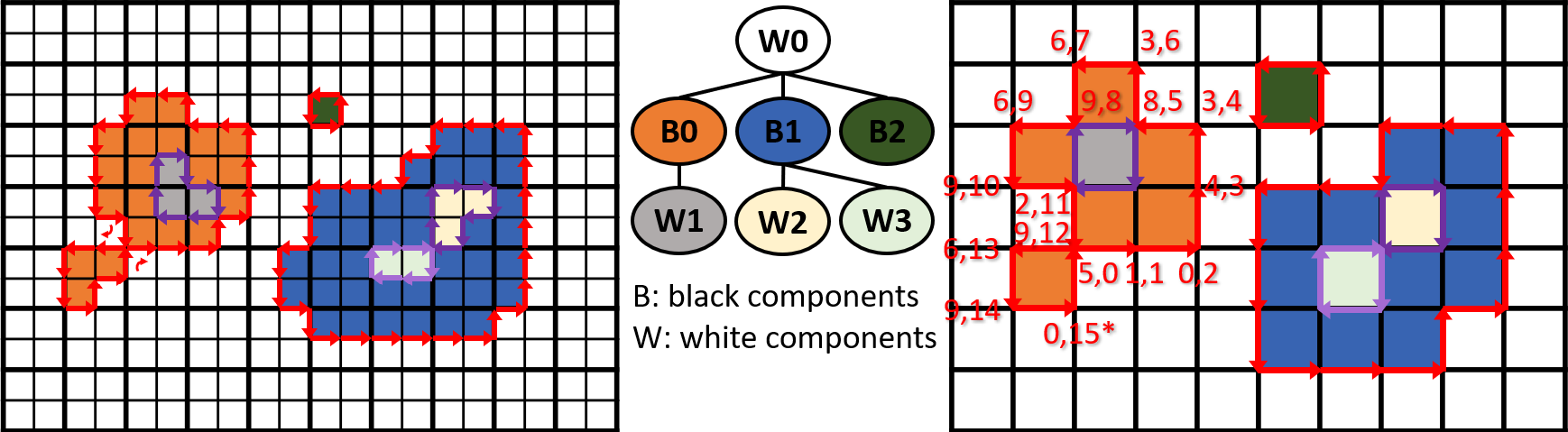}
  \caption{Left: An input binary image of 3 black components (orange, blue, and green) and 4 white components (white, grey, goose-yellow, and light green). Middle: Its region adjacency graph (RAG). Right: a topologically correct downsampled solution with the same RAG. We draw the outer and inner boundaries of black components in red and purples, respectively. Observe that a black components' outer boundary can self-intersect. For inner boundaries, one cannot intersect itself but different ones can intersect with each other. We also show the type-index and distance of each corner on the first component's outer boundary. The corner with the "last" flag on is asterisked.}
  \label{fig:method0}
\end{figure}

We can instead remove all the black quads and the resulting dangling vertices and edges from the binary image, and have a {\em white-quad mesh} of the binary image. However, different to the rule for black quads, we consider two white quads to be neighbors if and only if they share an edge. In other words, {\em black quads use the 8-neighborhood rule but white quads use the 4-neighborhood rule}. With the black-quad mesh and white-quad mesh of a binary image both built, we can now partition the whole quad mesh into black-quad connected components ("black components" in short) and white components that together fully cover the whole quad mesh in a non-overlapping manner. We also build the {\em region adjacency graph (RAG)} of the binary image, which is a graph in which each node represents a component, and two nodes are connected by an edge if and only if the two components are adjacent (\cite{shaprio01}). Note that every pair of adjacent components must be one black and one white. See Figure~\ref{fig:method0} for an example.


We now introduce the key concept behind our topology-preserving constraint formulation - the {\em boundaries} of connected components. We use notions of standard half-edge mesh data structures~\cite{CGAL} .
\begin{definition}
A boundary of a connected component's is a closed loop of consecutive half-edges such that every such half-edge's face belong to the component and its opposite half-edge's face does not not belong to the component.
\end{definition}
A boundary is {\em outer} if it is not of a hole. Otherwise, it is an {\em inner} boundary. Note that the consecutive half-edges of a outer and inner boundary go in the counter-clockwise and clockwise directions, respectively. 

The case of the half-edges on the boundary of the whole quad mesh creates some confusion (because the face pointers of the opposites of these half-edges are null). To simplify discussions, we assume the faces on the boundary of the whole quad mesh are all white, and ignore the outer boundary of the white component that touches the quad mesh boundary (e.g.,  Figure~\ref{fig:method0}). To support cases where there exist black faces touching the binary image boundary, we temporarily add a ring of white faces along the image boundary and later crop out the resulting white big-pixels. We now show an important lemma:
\begin{lemma}
\label{lem:boundary}
Faces of the opposite of all the half-edges in a boundary all belong to the same component.
\end{lemma}
We show the proof in Supplementary Materials.



Lemma~\ref{lem:boundary} implies that for every pair of adjacent black and white component as defined in the RAG of the original binary image, there exist exactly one closed loop of consecutive half-edges in counter-clockwise order (which is the outer boundary of the black component) that separates them.


\subsection{Optimization Problem Formulation}


We now formulate an integer programming (IP) optimization problem such that every solution constitutes a topologically correct downsampling result. We found that simply denoting the black-or-white decision of every big-pixels by a Boolean variable cannot cope with the complexity of the problem. Instead, we have to encode the {\em per-big pixel coverage by each components}. We first denote all components in the original binary image as $C_i$, $0 \le i < N_c$, $N_c$ is the number of components (both black and white). Now, for each component $C_i$, we collect all big-pixels that can potentially cover its pixels. The default way is that a pixel can be covered by a big-pixel if it is within the $A$-by-$B$ block of pixels of the big-pixel. We extend this idea to allow a big-pixel to cover its {\em nearby} pixels outside its block by a distance threshold in the original resolution. Specially, we say a big-pixel at coordinate $(X,Y)$ in the downsampled image, $0 \le X < W/A$ and $0 \le Y < H/B$, can cover a pixel at coordinate $(x,y)$ in the original image if:
\begin{align}
X*A -dx \le x \le ((X+1)*A - 1) + dx, \nonumber\\
Y*B -dy \le y \le ((Y+1)*B - 1) + dy.
\label{equ:coverage}
\end{align}
$dx$ and $dy$ are distance thresholds in the two directions. We use $\lfloor A/4 \rfloor$ for $dx$ and $\lfloor B/4 \rfloor$ for $dy$ in our implementation.

\noindent\textbf{Boolean Variables Declaration.} We denote the collected set of big-pixels that can cover component $C_i$ as:
\begin{equation}
P^i_j,
\label{equ:bigpixel}
\end{equation}
$0 \le j < N_i$, $N_i$ is the number of big-pixels that can cover $C_i$. For every $P^i_j$, we create a Boolean variable of the same name. We say $P^i_j$ is the $j$-th covering big-pixel candidate for $C_i$. Note that a big-pixel can be a covering candidate for multiple components. We now formulate the four sets of constraints for the IP problem as follows.

\noindent\textbf{1) Full and Non-Overlapping Coverage Constraint.} Every big-pixel has to be covered by exactly one component. This means:
\begin{equation}
\forall_{(X,Y)} \; \sum_{i,j}{P^i_j(X,Y)} = 1.
\end{equation}
$P^i_j(X,Y)$ denotes any $P^i_j$ at coordinate $(X,Y)$ in the downsampled binary image. 

\noindent\textbf{2) Component Non-Emptiness Constraint.} Every component has to be covered by at least one big-pixel otherwise it would disappear. That is:
\begin{equation}
\forall_i \; \sum^{N_i}_j{P^i_j} \ge 1.
\end{equation}

\noindent\textbf{3) Local Neighborhood Correctness Constraint.} Topological correctness in the neighborhood of each big-pixel is guaranteed by ensuring that two different black components can not be adjacent to each other by the 8-neighborhood rule and two different white components cannot be adjacent to each other by the 4-neighborhood rule. Accordingly, we collect all pairs of mutually incompatible big-pixel candidate Boolean variables (for example, two adjacent big-pixel candidates that belong to different black or white components) as $(P^m_0,P^m_1)$, $0 \le m < N_m$, $N_m$ is the number of such pairs. We then have:
\begin{equation}
\forall_{m} \; (P^m_0 + P^m_1) \le 1.
\label{equ:incompatible}
\end{equation}


Next, we introduce a constraint to ensure that, for every pair of adjacent black and white components as defined in the original RAG, there exists {\em exactly one closed loop of consecutive half-edges that separate the two components}.

\begin{figure}[t]
  \centering
  \includegraphics[width=1\linewidth]{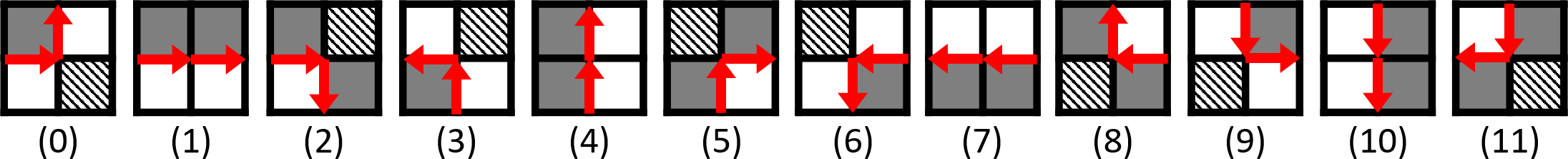}
  \caption{The 12 possible corner configurations and their type-indices. Each has one incoming half-edge and one outgoing half-edge, one or two left hand-side (LHS) big-pixels (black), and one or two right hand-side (RHS) big-pixels (white) defined. Each configuration has a "pointing-to" position defined, e.g., for the 0-th it is on the top.}
  \label{fig:method1}
\end{figure}

\noindent\textbf{4) Boundary-Preservation Constraint.} Our main idea is to encode a solved boundary as one closed loop of consecutive {\em corners} in a downsampled binary image. A corner is two consecutive half-edges, $v0$-to-$v1$ and $v1$-to-$v2$, that intersect at $v1$. There exist exactly 12 possible {\em corner configurations}, as shown in Figure~\ref{fig:method1}. Note that up to two corners can lie on the same vertex in a non-overlapping nor crossing-over manner (e.g., 5-th can collocate with 11-th, but not 4-th).

For every boundary between a black component $Cb$ and a white component $Cw$, correctness in terms of local connectivity is ensured as follows. Observe that each corner configuration has one or two adjacent "left hand-side (LHS)" big-pixels and one or two "right hand-side (RHS)" big-pixels defined (see Figure~\ref{fig:method1}). We then require that a corner is part of a boundary {\em if and only if all its RHS big-pixels exist in $Cb$ and all its LHS big-pixels exist in $Cw$}. This is realized by adding the following constraints for every possible corner candidate, momently denoted as a Boolean variable $Corner$:
\begin{equation}
0 \le (\sum{P^{RHS}_{Cb}} + \sum{P^{LHS}_{Cw}}) - N * Corner\le N-1.
\label{equ:corner}
\end{equation}
$P^{RHS}_{Cb}$ are all possible big-pixel Boolean variables belong to black component $Cb$ at the corner's RHS side, and $P^{LHS}_{Cw}$ are all possible big-pixel Boolean variables belong to white component $Cw$ at the corner's LHS side. $N$ equals to sum of the number of all possible $P^{RHS}_{Cb}$ and the number of all possible $P^{LHS}_{Cw}$. 

In practice, for every boundary, we exhaustively search all 12 possible corner configurations on every possible inner vertices of the downsampled binary image, and enumerate all possible "corner candidates" as Boolean variables for the boundary as the ones that have both non-empty $P^{RHS}_{Cb}$ and $P^{LHS}_{Cw}$ sets. Equations~\ref{equ:corner} are then applied to every enumerated corner candidates of the boundary. 

However, the above constraints still don't guarantee that there exists only {\em one} close loop of corners. Inspired by the "no-island" constraint proposed in~\cite{10.1145/2897824.2925935}, we further create a "distance" integer variable and a "last" Boolean variable to pair with every enumerated corner candidates of a boundary. We denote the former as $Dist_k$ and the latter as $Last_k$ for corner candidate variable $Corner_k$, $0 \le k < N^{cc}$, $N^{cc}$ is the number of corner candidates of the boundary. Here, our goal is to enforce that {\em distance values of consecutive corners along a boundary must be monotonically increasing}, with exactly one exception at the corner with the "last" flag on. This effectively ruled out any possibly of having multiple closed loops as each loop needs at least one "last" flag to be true. We now have:
\begin{align}
\forall_k \; (Corner_k - ((\sum_l{Dist_l}) - Dist_k) - Last_k * \text{BIG}) \le 0,
\end{align}
where $Dist_l$ are distance value variables of compatible corners at $Corner_k$'s pointing-to position. $\text{BIG}$ is a per-boundary integer constant that is guaranteed to be bigger than the largest possible length of the downsampled boundary (e.g., set to the length of the boundary in the original binary image). We also require that a corner's distance value must be zero if the corner itself is false:
\begin{equation}
\forall_k \; Dist_k \le BIG * Corner_k.
\end{equation}
Finally, we require that every boundary has exactly one "last" flag set to true:
\begin{equation}
\sum_{k \in \text{Boundary}}{Last_k} = 1.
\end{equation}
See Figure~\ref{fig:method0} right for an example of a solved boundary. 

In summary, the above four sets of constraints have ensured that a solved binary image, which is the summation of all the active big-pixel variables, shall have the same RAG as the original. A proof is in the Supplementary Materials. Finally, we design our cost function as follows.

\begin{figure}[t]
\begin{minipage}{0.24\textwidth}
\includegraphics[width=\linewidth]{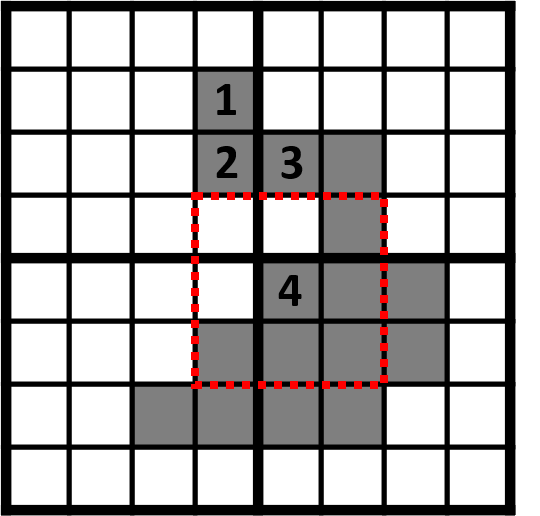}
\end{minipage}
\begin{minipage}{0.74\textwidth}
\caption{\label{fig:score}We downsample a binary image of one black component and one white component by factor 4 in width and height. For the upper-left big-pixel of the black component, it's score value is 16, which is contributed by the four numbered black pixels with scores 6, 6, 3, and 1, respectively. To see this, the $4$-th pixel's window is shown in red. It overlaps with the big-pixel at one pixel, therefore contributing 1 to the big-pixel's score.}
\end{minipage}


\end{figure}

\noindent\textbf{Objective Function.} Our goal is to assign every big-pixel candidates of every component (i.e., $P^i_j$ as defined in Equation~\ref{equ:bigpixel}) a "score" value, $S^i_j$, and formulate the objective function as to maximize the sum of scores of active big-pixels:
\begin{equation}
\max_{P^i_j} \sum_{i,j} (S^i_j * P^i_j),
\label{equ:objective}
\end{equation}
where $i$ denote the component index and $j$ denotes the big-pixel candidate index in a component. Our general idea is to raise a score whenever an original pixel of a particular component is presented by an active big-pixel of the same component, and the raise is inversely proportional to the distance in between. We propose a way to achieve this as follows. For every pixel of component $C_i$ at coordinate $(x,y)$, we enumerate all its nearby pixel coordinates within a $(2*dx+1)$-by-$(2*dy+1)$ window ($dx$ and $dy$ are coverage distance thresholds used in Equation~\ref{equ:coverage}). Next, for every pixel in the window, we accumulate $1$ to the score of the corresponding big-pixel candidate. See Figure~\ref{fig:score} for an example. 

We provide an analysis on the all possible outcomes of solving the IP problems in the Supplementary Materials.





\section{A Baseline Dilation-Based Method}
\label{sec:dilation}

\begin{figure}[t]
  \centering
  \includegraphics[width=1\linewidth]{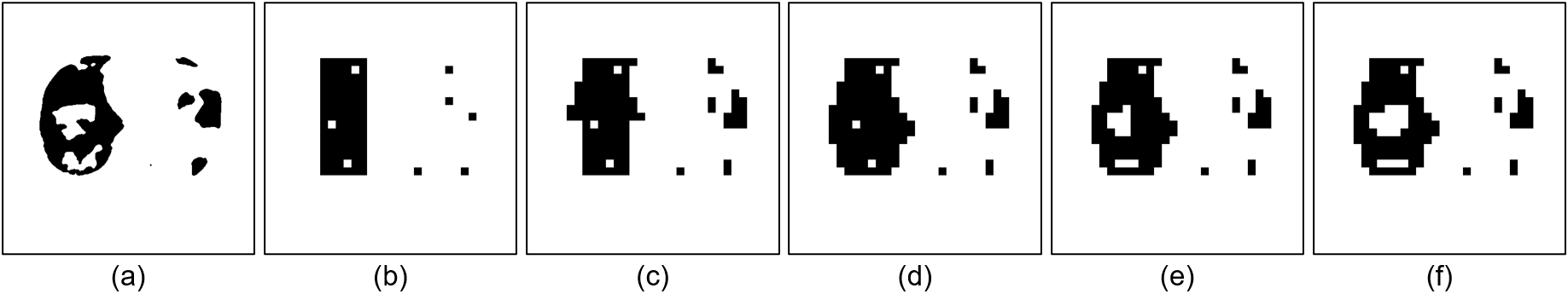}
  \caption{Our baseline dilation-based method. (a) An input binary image. (b) The initial downsampled image. Each component is presented as a dot or a bounding box of inner holes. (c) After 10 iterations of black component dilation. (d) After black component dilation converged. (e) After 5 iterations of white component dilation. (f) Final result.}
  \label{fig:dilation}
\end{figure}

In short, we first enumerate all the black and white components in the input binary image. Next, in the downsampled image (initialized as all white), for every component, we create a trivial and topologically correct representation of it. We do this by simply creating a dot at the center for every component. For components with holes, we create a bounding box of inner dots. Next, we iteratively {\em dilate} first all the black components then all the white components one pixel by one pixel. It becomes trivial to preserve topology for each dilation (i.e., skip ones that break or merge components). We stop until no more dilation operations can be found, or a time limit is reached. See Figure~\ref{fig:dilation} for an example.

\section{Results and Applications}
\label{sec:result}

We implemented our method in C++ and used Gurobi 11.0~\cite{gurobi} to solve the IP problems. We tested on a Windows PC with AMD 7 Ryzen 5800X CPU, 32GB RAMs, and NVIDIA RTX 3060 GPU. In Section~\ref{sec:result_medical}, we tested our method versus other downsampling methods on the segmentation masks of a medical image dataset. We show results on more kinds of segmentation masks in the Supplementary Materials. In Section~\ref{sec:result_speed}, we show that the speeds of two important binary image operations - persistent homology (PH) computation and shortest path calculation, can be significantly improved with small impact to the accuracy via straightforward applications of our method. 

We re-implemented the ACN~\cite{decenciere2007adaptive} method in C++ as the authors did not provide codes. Since ACN only does factor-2 downsampling, we run it multiple times to achieve downsampling at factors greater than 2. We use IoU and Dice scores of the foreground (black) components to measure pixel-wise similarity between two binary images. We use \textbf{Betti number error}, which measures sum of absolute errors of the zeroth and first Betti numbers, and \textbf{PH distance}, which measures the bottleneck distances of persistent diagrams~\cite{maria2014gudhi} (computed by~\cite{giotto-tda}), to measure topological similarities.

\subsection{Medical Image Segmentation Masks Downsampling}
\label{sec:result_medical}

We have found that visualizing the segmentation masks in medical datasets in reduced resolutions are helpful for users to identify intricate details such as tiny components or holes, and components that are very close-by but actually are separated. This is because small details are now represented by bigger pixels. Inspired by this, we tested several downsampling methods on the 542 segmentation masks (resolution 512x512) of lung coronavirus in the China National Center for Bioinformation (CNCB) dataset~\cite{CNCB_dataset}. We show quantitative results in terms of topological accuracy, pixel-wise similarity, and speed in Table~\ref{tab:medical_result}. We see that our method generated completely topologically correct results. Our results also have the lowest PH distances to the original among all methods. Remarkably, our results also achieved nearly the same levels of pixel-wise similarity as the non topology-preserving methods. The baseline dilation-based method also produced completely topologically correct results, but the pixel-wise accuracy metrics are much worse. Note that the computational costs of our methods are inversely proportional to the downsampling rates - this is because the higher the downsampling the fewer the big-pixel candidates. We consider our method's speed to be usable in interactive applications (less than 1 seconds for factor $\ge 4$ cases). Our method cannot find feasible solutions for a few cases of the masks - at downsampling factor 2, 4, 8, and 16, there are 2, 1, 3, and 8 infeasible cases, respectively. Infeasible cases are generally detected instantly by our IP solver (times are included in our average time computations).

\begin{figure}[t]
	\begin{minipage}{1\linewidth}
		\vspace{2pt}
		\centerline{\includegraphics[width=\textwidth]{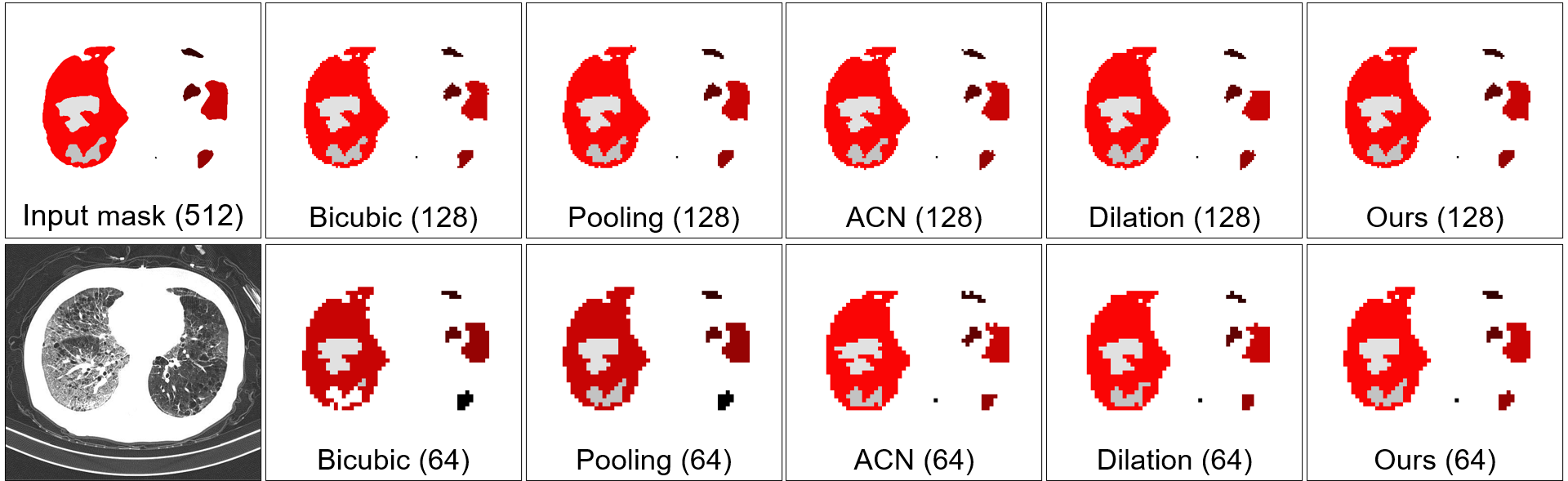}}
			\vspace{2pt}
		\centerline{\includegraphics[width=\textwidth]{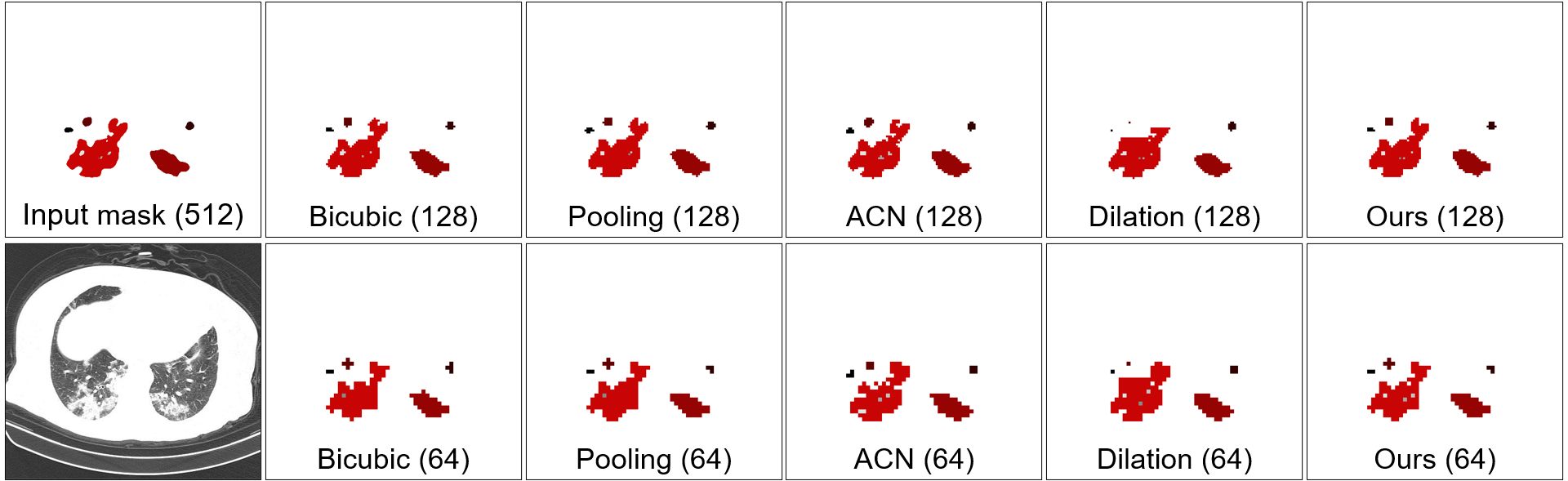}}
			\vspace{2pt}
		\centerline{\includegraphics[width=\textwidth]{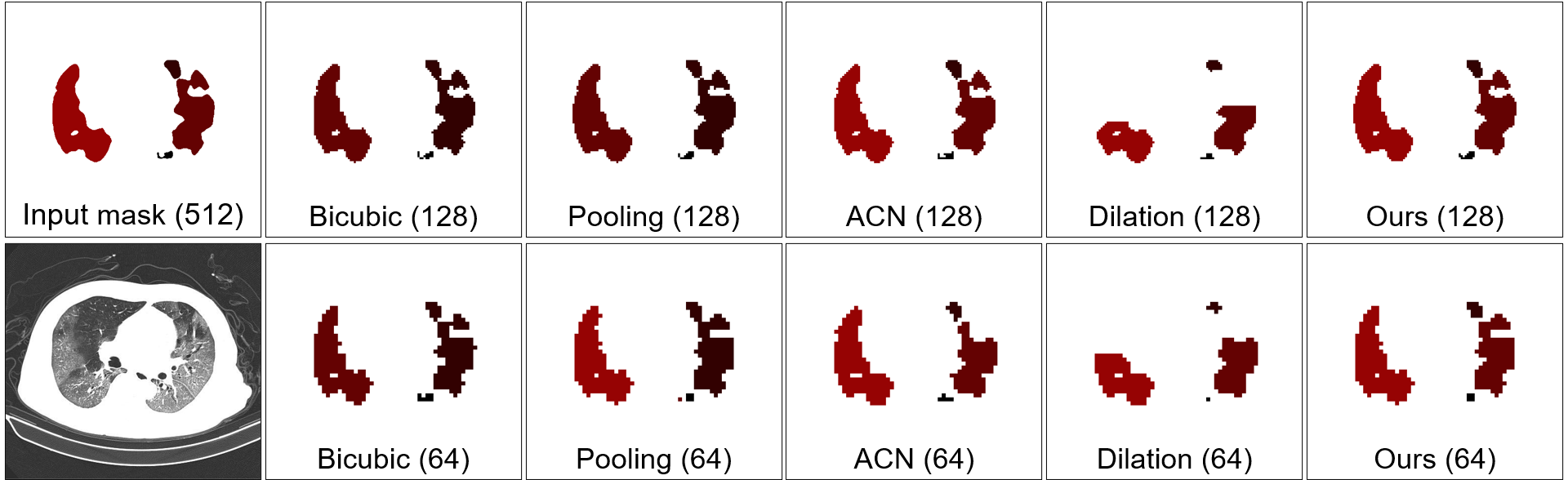}}
	\end{minipage}
	\caption{Visualization of segmentation masks in downsampled resolutions. Details such as tiny components and holes are made easier to observe with bigger pixels. We deploy a coloring scheme to highlight the fact that non topology-preserving methods may produce results with erased small components/holes and merged/broken components.}
    \label{fig:medical_result}
\end{figure}

We show side-by-side qualitative comparisons in Figure~\ref{fig:medical_result}. As a mean for visualization, we show different foreground and background components in different colors (levels of red and white/grey for foreground and background components, respectively). This coloring scheme helps to highlight an important fact that non topology-preserving downsampling methods may merge separate components (or holes) into one, or break one component or hole into multiple ones.


\begin{table}
\scriptsize
\begin{center}
\caption{\label{tab:medical_result}Quantitative and speed comparisons of different downsampling methods on the 542 segmentation masks in the CNCB dataset~\cite{CNCB_dataset} to different sizes. {\color{red}Best} and {\color{blue}second-best} results are marked in red and blue, respectively.}
\label{table:headings}
\begin{tabular}{l|ccccc|ccccc}

 & \multicolumn{5}{c|}{512x512 to 256x256 (factor 2)} & \multicolumn{5}{c}{512x512 to 128x128 (factor 4)} \\

\hline
Method & $\uparrow$ IoU & $\uparrow$ Dice & \makecell{$\downarrow$ Betti \\ num. error} &  \makecell{$\downarrow$ PH \\ distance} & \makecell{$\downarrow$ Avg. \\ time (s)} & $\uparrow$ IoU & $\uparrow$ Dice & \makecell{$\downarrow$ Betti \\ num. error} &  \makecell{$\downarrow$ PH \\ distance} & \makecell{$\downarrow$ Avg. \\ time (s)}\\
\hline
Bicubic  & 93.08\% & \textcolor{blue}{96.34\%} & 0.061 & 0.018 & \textcolor{red}{0.002} & \textcolor{blue}{85.54\%} & \textcolor{blue}{91.78\%} & \textcolor{blue}{0.133} & 0.034 & \textcolor{red}{0.002}\\
Pooling  & \textcolor{red}{93.41\%} & \textcolor{red}{96.53\%} & \textcolor{blue}{0.046} & \textcolor{blue}{0.015} & 0.615 &  \textcolor{red}{85.99\%} & \textcolor{red}{92.03\%} & 0.144 & 0.046 & \textcolor{blue}{0.160}\\
ACN~\cite{decenciere2007adaptive}  & 91.78\% & 95.61\% & 0.092 & 0.051 & \textcolor{blue}{0.161} & 78.78\% & 87.52\% & 0.135 & 0.085 & 0.256\\
Dilation & 59.71\% & 73.78\% & \textcolor{red}{0} & 0.026 & 2.348 & 61.69\% & 75.39\% & \textcolor{red}{0} & \textcolor{blue}{0.028} & 0.544\\
Ours  & \textcolor{blue}{93.10\%} & \textcolor{blue}{96.34\%} & \textcolor{red}{0} & \textcolor{red}{0.005} & 1.976 & 85.12\% & 91.55\% & \textcolor{red}{0} & \textcolor{red}{0.012} & 0.819\\
\hline\noalign{\medskip}
& \multicolumn{5}{c|}{512x512 to 64x64 (factor 8)} & \multicolumn{5}{c}{512x512 to 32x32 (factor 16)} \\
\hline
Method & $\uparrow$ IoU & $\uparrow$ Dice & \makecell{$\downarrow$ Betti \\ num. error} &  \makecell{$\downarrow$ PH \\ distance} & \makecell{$\downarrow$ Avg. \\ time (s)} & $\uparrow$ IoU & $\uparrow$ Dice & \makecell{$\downarrow$ Betti \\ num. error} &  \makecell{$\downarrow$ PH \\ distance} & \makecell{$\downarrow$ Avg. \\ time (s)}\\
\hline
Bicubic  & 73.27\% & \textcolor{blue}{83.46\%} & \textcolor{blue}{0.339} & 0.091 & \textcolor{red}{0.002} & \textcolor{red}{55.37\%} & \textcolor{blue}{68.28\%} & 0.779 & 0.184 & \textcolor{red}{0.002}\\
Pooling  & \textcolor{red}{73.33\%} & 83.21\% & 0.404 & 0.115 & \textcolor{blue}{0.046} &  54.35\% & 66.75\% & 0.969 & 0.214 & \textcolor{blue}{0.017}\\
ACN  & 60.81\% & 74.37\% & 0.356 & 0.150 & 0.335 & 41.00\% & 56.42\% & \textcolor{blue}{0.600} & 0.181 & 0.414\\
Dilation & 53.40\% & 68.07\% & \textcolor{red}{0} & \textcolor{blue}{0.051} & 0.137 & 35.96\% & 50.49\% & \textcolor{red}{0} & \textcolor{blue}{0.085} & 0.043\\
Ours  & \textcolor{blue}{73.31\%} & \textcolor{red}{83.55\%} & \textcolor{red}{0} & \textcolor{red}{0.030} & 0.612 & \textcolor{blue}{54.85\%} & \textcolor{red}{68.92\%} & \textcolor{red}{0} & \textcolor{red}{0.062} & 0.563\\
\hline
\end{tabular}

\end{center}
\end{table}

\begin{table}
    \centering
    \caption{\label{tab:PH}Average times and PH distances of PH computations at different downsampled resolutions, tested on the CNCB dataset.}
    \begin{tabular}{c||c|c|c|c|c}
         Image sizes: &  512 (original) & 256 (factor-2) & 128 (factor-4) & 64 (factor-8) & 32 (factor-16) \\
         \hline
         Avg. time (s) & 0.965 & 2.157 & 0.864 & 0.624 & 0.568 \\
         PH distance & - & 0.005 & 0.012 & 0.030 & 0.062 \\
    \end{tabular}
\end{table}


\subsection{Binary Image Operations Speed Improvement}
\label{sec:result_speed}

\textbf{Persistent Homology (PH) Computation Speed-up.} As shown in Table~\ref{tab:medical_result}, the quantitative differences of persistent diagrams (i.e., PH distances) between our downsampled results and the original are in general very small. A visual comparison of persistent diagrams, shown in the Supplementary Materials, also confirmed this. Therefore, one possible approach to speed up PH computation is by substituting the original binary images by smaller, downsampled versions computed by our method. For this approach to work, the time needed by our method to compute a downsampled image has to be shorter than the time saved by computing PH on a smaller (downsampled) image than the original. Our timing comparisons (shown in Table~\ref{tab:PH}) using PH computation done by a commonly used tool~\cite{giotto-tda} confirmed that this is the case when downsampling factor is $\ge 4$. 


\begin{table}[t]
    \scriptsize
    \centering
    \caption{\label{tab:shortest_path} In average, conducting 200 Dijkstra shortest path computations with random start and end pixels needs about 37.67 seconds in a 512x512 binary image. We demonstrate that doing so in downsampled images can have significant time saving with small impact to the accuracy. We show the average shortest distance errors, numbers of false positives (FP) and false negatives (FN), and total time (200 computations) using different downsampling methods at different sizes. Only using our method and the dilation method is free of FP and FN cases, while ours has smaller distance errors.}

    \begin{tabular}{c|c|c|c|}
         Avg. dist. error, FP, FN| time(s) &  128 (factor-4) & 64 (factor-8) & 32 (factor-16)\\
         \hline
         Bicubic & 1.65, 0.00, 0.78| 2.49 & 5.05, 0.00, 2.12| 0.95 & 11.57, 0.00, 1.85| 0.96 \\
         Pooling & 2.63, 0.00, 0.00| 2.96 & 6.60, 0.01, 0.26| 1.21 & 14.64, 0.01, 0.40| 1.17 \\
         ACN~\cite{decenciere2007adaptive} & 1.80, 0.01, 0.10| 3.22 & 5.82, 0.00, 0.29| 1.92 & 13.44, 0.05, 0.47| 2.00 \\
         Dilation & 0.38, 0.00, 0.00| 3.66 & 1.00, 0.00, 0.00| 1.48 & \hphantom{1}2.52, 0.00, 0.00| 1.27 \\
         Ours & 0.17, 0.00, 0.00| 3.63 & 0.19, 0.00, 0.00| 1.34 & \hphantom{1}0.47, 0.00, 0.00| 0.99 \\
    \end{tabular}

    
\end{table}

\noindent\textbf{Shortest Path Speed-up.} Dijkstra shortest path computation in the black components is an operation that may take place if a binary image is taken as a geographical map (e.g., black components as land masses) in game design. We can speed up the computation by leveraging binary image downsampling methods. Our approach is as follows. Assume we have an original image, $A$, and an downsampled image, $B$. Our goal is to calculate a shortest path from pixel $p0$ to pixel $p1$ (both black pixels) in $A$. We now find the {\em corresponding} big-pixels of $p0$ and $p1$ in $B$, denoted as $P0$ and $P1$. This is done by: 1) identify the pixel $p$'s component index. 2) Check if the big-pixel at $p$'s corresponding coordinate in $B$ is belong to the same component. 3) If yes, use the big-pixel. 4) If not, find the closest big-pixel (to big-pixel boundaries in the original resolution) with the same component index. We then take the computed shortest path from $P0$ to $P1$ in B, scaled back to the original resolution, as the approximated shortest path for $p0$ to $p1$. Note that in this application, using non topologically-correct downsampled images is unsuitable: it is because when accurate component indices of big-pixels are unavailable, we can only find each pixel's corresponding big-pixels approximately by distance. This may cause originally unreachable pairs of points (i.e., in different components) to become reachable in the downsampled image ("false positive" cases) and originally reachable pairs of points to become unreachable ("false negative" cases). 

We conduct experiments by randomly choosing 200 pairs of starting and ending pixels in each of the 542 binary images in the CNCB dataset. We use Boost Graph Library~\cite{boost_graph}'s Dijkstra implementation. Note that we allow choosing two pixels in different black components (no paths in between). We show testing results in Table~\ref{tab:shortest_path}. Note that time saving happens because typically shortest path computation are done many times in a binary image, and total time saving will be greater than the time spent on doing downsampling once per image.


\section{Conclusion and Future Work}
\label{sec:conclusion}

In summary, our method generates downsampled binary images that are guaranteed to have the same RAG as the original. This is in fact a stronger guarantee then just having the same Betti numbers - as two 2D graphs with the same RAG must have the same Betti numbers, but the inverse is not necessarily true (e.g., same number of holes but appear in different connected components). 

A main limitation is to handle binary images with very thin structures such as road networks and blood vessels. Currently, such structures may not be satisfactorily preserved because our objective function design preserves black and white components with the same preference. Therefore, an interesting research direction may be to design alternative cost functions to handle such binary images. Codes are available at: github.com/pengchihan/BinaryImageDownsampling.



\textbf{Acknowledgements.} This work is funded by the National Science and Technology Council of Taiwan (project number 111R10286C).

%
%
\bibliographystyle{splncs04}
\bibliography{main}

\clearpage
\section{Supplementary Materials}

\subsection{Aerial Image Dataset Roof Masks Downsampling Results}
\label{sec:aerial}

\begin{figure}[t]
	\begin{minipage}{1\linewidth}
		\vspace{2pt}
		\centerline{\includegraphics[width=\textwidth]{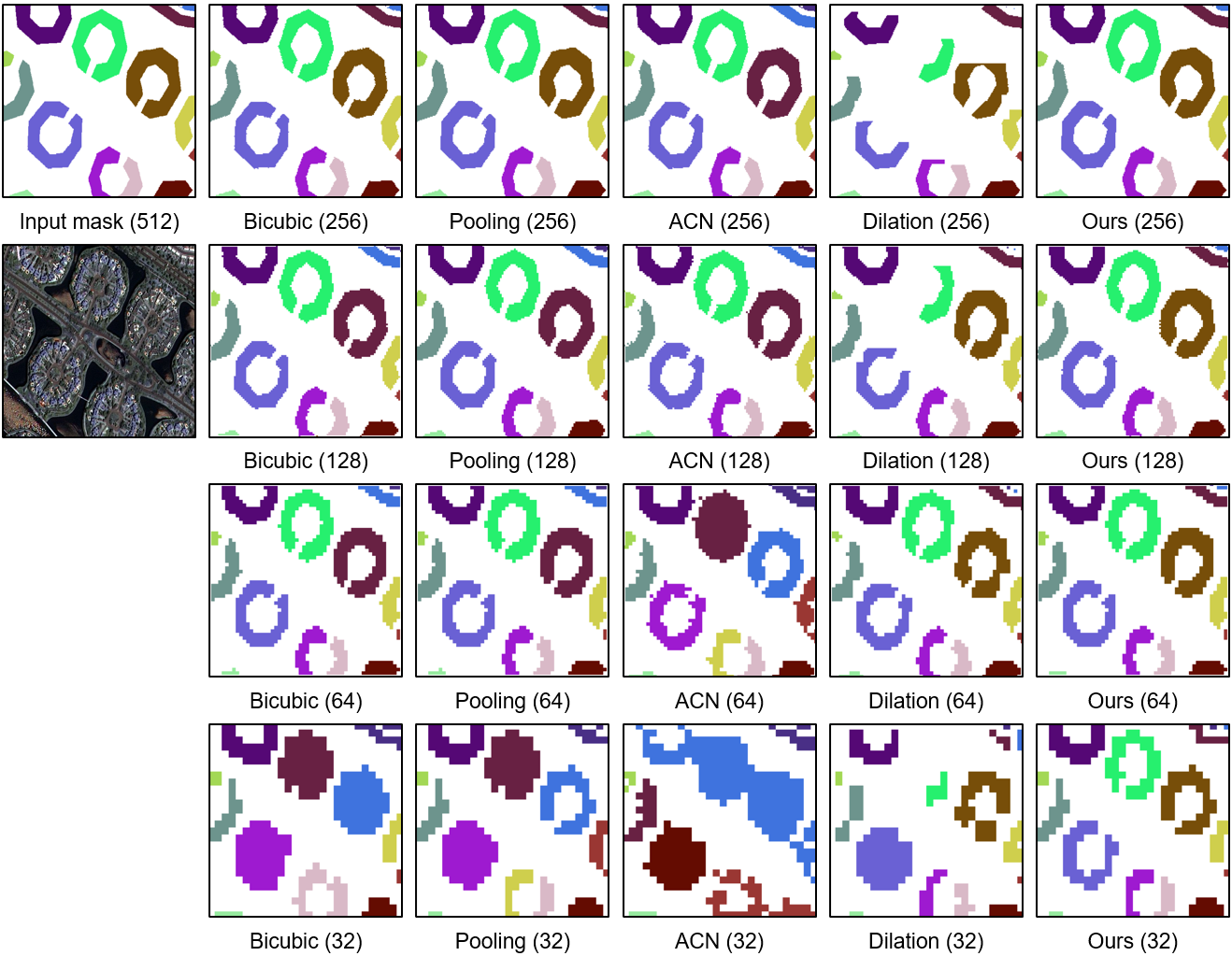}}
			\vspace{2pt}
		\centerline{\includegraphics[width=\textwidth]{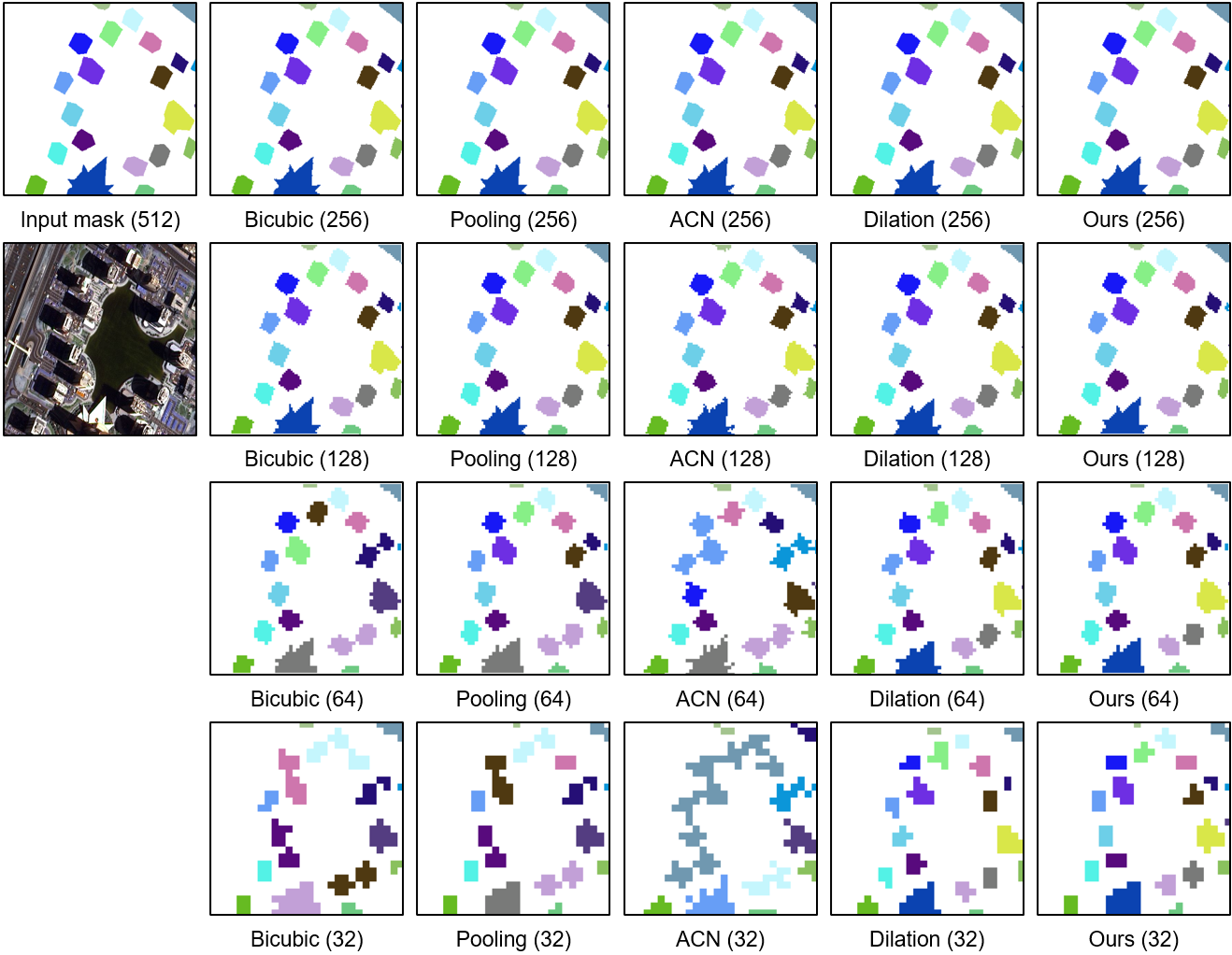}}
	\end{minipage}
	\caption{Aerial roof mask dataset downsampling results. These are challenging problems that non topology-preserving methods often generate results with altered topology.}
    \label{fig:aerial_result}
\end{figure}

We also tested on another binary image dataset consisting of building roof masks in real-world aerial images of urban areas (see Table~\ref{tab:aerial_result} and Figure~\ref{fig:aerial_result}). We built our dataset of 255 512x512 images by randomly subtracting sub-images from an satellites aerial image dataset (by Humans in the Loop group with the Mohammed Bin Rashid Space Center\footnote{https://humansintheloop.org/resources/datasets/semantic-segmentation-dataset-2/}), which originally contains 72 images of various sizes (from 509x544 to 2149x1479). As mentioned in Section~\ref{sec:method}, we temporarily added a ring of white pixels to the image boundaries to simplify boundary handling cases. We point out that this dataset is more challenging than the CNCB dataset we tested in Section~\ref{sec:result}, as it contains a lot of small and relatively thin components. This leads to even more topologically incorrect downsampling results (compared to CNCB dataset) by the traditional non topology-preserving methods. However, our IP-based method and the dilation-based method still produced completely topologically correct results. Our method's results still have similar levels of pixel-wise accuracy as the traditional methods, while the dilation method's accuracy is much worse. On the downside, this dataset is more challenging to have topologically correct results - there are 8, 3, 14, and 42 infeasible cases for downsampling factors 2, 4, 8, and 16, respectively.


\begin{table}
\scriptsize
\begin{center}
\caption{\label{tab:aerial_result}Quantitative and speed comparisons of different downsampling methods on the 255 segmentation masks in the aerial dataset to different sizes. {\color{red}Best} and {\color{blue}second-best} results are marked in red and blue, respectively.}
\label{table:aerial}
\resizebox{\textwidth}{!}{
\begin{tabular}{l|ccccc|ccccc}

 & \multicolumn{5}{c|}{512x512 to 256x256 (factor 2)} & \multicolumn{5}{c}{512x512 to 128x128 (factor 4)} \\

\hline
Method & $\uparrow$ IoU & $\uparrow$ Dice & \makecell{$\downarrow$ Betti \\ num. error} &  \makecell{$\downarrow$ PH \\ distance} & \makecell{$\downarrow$ Avg. \\ time (s)} & $\uparrow$ IoU & $\uparrow$ Dice & \makecell{$\downarrow$ Betti \\ num. error} &  \makecell{$\downarrow$ PH \\ distance} & \makecell{$\downarrow$ Avg. \\ time (s)}\\
\hline
Bicubic & 94.56\% & 97.15\% & 2.47 & 0.077 & \hphantom{1}\textcolor{red}{0.001} & 86.75\% & 92.50\% & 4.62 & 0.135 & \textcolor{red}{0.001}\\
Pooling & \textcolor{red}{94.81\%} & \textcolor{red}{97.29\%} & \textcolor{blue}{0.98} & \textcolor{blue}{0.056} & \hphantom{1}0.675 & \textcolor{red}{87.85\%} & \textcolor{red}{93.32\%} & \textcolor{blue}{2.43} & \textcolor{blue}{0.124} & \textcolor{blue}{0.174}\\
ACN~\cite{decenciere2007adaptive}& 93.59\% & 96.63\% & 5.15 & 0.215 & \hphantom{1}\textcolor{blue}{0.190} & 81.98\% & 89.74\% & 5.67 & 0.240 & 0.285\\
Dilation & 55.42\% & 67.81\% & \textcolor{red}{0.00}  & 0.174 & 37.702 & 54.41\% & 67.68\% & \textcolor{red}{0.00}  & 0.176 & 8.958\\
Ours & \textcolor{blue}{94.71\%} & \textcolor{blue}{97.24\%} & \textcolor{red}{0.00} & \textcolor{red}{0.031} & \hphantom{1}3.137 & \textcolor{blue}{87.33\%} & \textcolor{blue}{92.97\%} & \textcolor{red}{0.00}  & \textcolor{red}{0.055} & 3.180\\
\hline\noalign{\medskip}
& \multicolumn{5}{c|}{512x512 to 64x64 (factor 8)} & \multicolumn{5}{c}{512x512 to 32x32 (factor 16)} \\
\hline
Method & $\uparrow$ IoU & $\uparrow$ Dice & \makecell{$\downarrow$ Betti \\ num. error} &  \makecell{$\downarrow$ PH \\ distance} & \makecell{$\downarrow$ Avg. \\ time (s)} & $\uparrow$ IoU & $\uparrow$ Dice & \makecell{$\downarrow$ Betti \\ num. error} &  \makecell{$\downarrow$ PH \\ distance} & \makecell{$\downarrow$ Avg. \\ time (s)}\\
\hline
Bicubic & 78.45\% & 87.15\% & 10.14 & 0.213 & \textcolor{red}{0.001} & \textcolor{blue}{63.13\%} & \textcolor{blue}{75.97\%} & 13.57 & 0.292 & \textcolor{red}{0.001}\\
Pooling & \textcolor{red}{78.94\%} & \textcolor{red}{87.39\%} & \hphantom{1}\textcolor{blue}{7.28} & 0.195 & \textcolor{blue}{0.051}& \textcolor{red}{63.60\%} & \textcolor{red}{76.11\%} & \textcolor{blue}{11.65} & 0.279 & \textcolor{blue}{0.018}\\
ACN~\cite{decenciere2007adaptive}  & 66.77\% & 79.78\% & 10.57 & 0.249 & 0.365 & 48.28\% & 63.83\& & 12.84 & 0.287 & 0.445\\
Dilation & 48.23\% & 62.75\% & \hphantom{1}\textcolor{red}{0.00}  & \textcolor{blue}{0.174} & 2.296 & 34.18\% & 48.55\% & \hphantom{1}\textcolor{red}{0.00}  & \textcolor{blue}{0.194} & 0.496\\
Ours & \textcolor{blue}{78.62\%} & \textcolor{blue}{87.41\%} & \hphantom{1}\textcolor{red}{0.00}  & \textcolor{red}{0.083} & 3.494 & 61.37\% & 74.60\% & \hphantom{1}\textcolor{red}{0.00}  & \textcolor{red}{0.115} & 5.348\\
\hline
\end{tabular}
}
\end{center}
\end{table}

\clearpage


\subsection{Comparisons to works by Passat et al. and Ngo et al.~\cite{passat:hal-03630330},~\cite{Ngo2014}, and~\cite{6690186}}

\begin{table}
    \centering
    \caption{\label{sup:tab1} Comparing~\cite{passat:hal-03630330} and ours. We tested on both our datasets (CNCB and roof mask dataset) of 512x512 input images. A non-success for [1] means it cannot reach a feasible solution. For ours it means the solver declares the problem to be infeasible (with default coverage value) or could not finish in time. We find that [1] often cannot reach a feasible solution for downsampling tasks w/ factor $>2$. Moreover, [1]'s results' PH distances are much worse.}
    \begin{tabular}{c|c|c|c|c}
          & IOU, Dice & PH dist & success\% & time \\
         \hline
         $[1]$, factor2 & 94.07, 96.89 & 0.247 & 82.00\% & 1.08 \\ 
         Our, factor2 & 93.91, 96.79 & 0.018 & 98.00\% & 2.56 \\ 
         \hline
         $[1]$, factor4 & 85.64, 91.94 & 0.241 & 41.00\% & 2.35 \\ 
         Our, factor4  & 86.23, 92.26 & 0.034 & 99.00\% & 2.00 \\ 
         \hline
         $[1]$, factor8 & 71.18, 82.28 & 0.181 & 10.35\% & 4.81 \\ 
         Our, factor8  & 75.97, 85.48 & 0.057 & 96.95\% & 2.05 \\ 
         
    \end{tabular}
\end{table}

\cite{Ngo2014} and~\cite{6690186} both analyzed the necessary and sufficient conditions for a 2D binary image to retain topology after arbitrary rigid transformations (translations and rotations, but no scaling). In~\cite{6690186}, two algorithms to transform a 2D binary image to be a topology-invariant-under-affine-transformation one are proposed. The first is by fixing ”at-fault” pixels (ones that can break topology after rigid transforms) one-by-one and the second is by doing a super-resolution (but no subsequent downsampling is discussed). Note that down-scaling is not a rigid transformation so their results are not directly applicable to our problems. In~\cite{passat:hal-03630330}, the authors extend the discussions to affine transformations (including scaling) and proposed an algorithm to do arbitrary topology-invariant affine transformation. In short, the algorithm (Alg.3 in their paper) starts by subdividing the initial binary image to fit in the target (affine transformed) grid, with possibly lots of arbitrarily-shaped polygons. Next, the subdivided polygon mesh is converted to one that is fully compatible to the target grid by applying morphological operations (erosion and dilation) one polygon-by-one. The operations are picked in a greedy, gradient-descent manner, sorted by an elegant scoring function (their Eq.85). However, the algorithm doesn’t have the guarantee that a feasible solution (if one exists) is always reached (note that it starts with an infeasible solution). In this regard, we consider it less reliable than our baseline greedy dilation-based approach (Sec.~\ref{sec:dilation} in our paper) because ours instead start with a feasible solution and improves upon it by applying topology-preserving dilation operations iteratively. We got the source codes from the authors and fixed some issues to make it work for large downsampling tasks. See Table~\ref{sup:tab1} for testing results.

\subsection{Proof of Lemma~\ref{lem:boundary}}

We recap the Lemma here:

\textit{Faces of the opposite of all the half-edges in a boundary all belong to the same component.}

\begin{figure}[]
  \centering
  \includegraphics[width=0.85\linewidth]{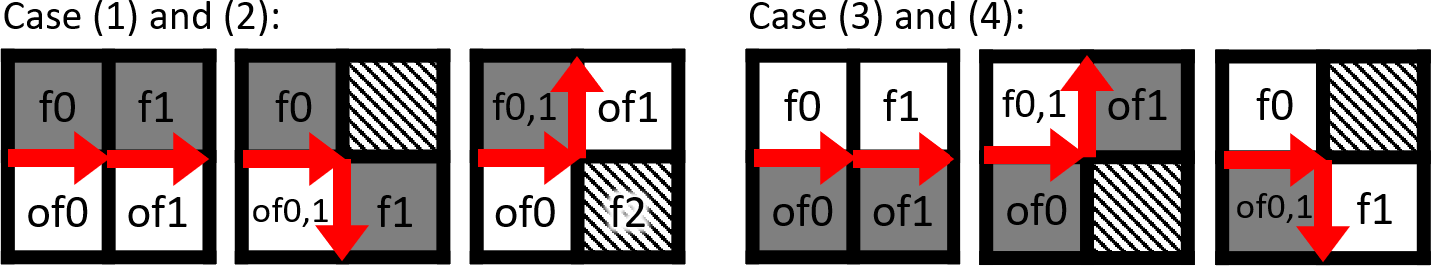}
  \caption{Showing the cases for the proof of Lemma~\ref{lem:boundary}.}
  \label{fig:proof_cases}
\end{figure}

\begin{proof}
We prove it by showing that the faces of the opposite of every two consecutive half-edges must belong to the same component. We denote two consecutive boundary half-edges as $e0$ and $e1$,their faces as $f0$ and $f1$, and their opposite's faces as $of0$ and $of1$, respectively. We discuss four possible cases of a boundary: (1) a outer boundary of a black component, (2) an inner boundary of a black boundary, (3) a outer boundary of a white component, and (4) an inner boundary of a white component. Illustrations are shown in Figure~\ref{fig:proof_cases}.

For case (1) and (2), if $f0$ and $f1$ share an edge, then $of0$ and $of1$ must belong to the same white component because they share one edge. If $f0$ and $f1$ shares only a vertex, then $of0$ and $of1$ are actually the same face. Now for the case that $f0$ and $f1$ are the same face, assume for a moment that $of0$ and $of1$ belong to different white components. This means that the face, call it $f2$, that is adjacent to both $of0$ and $of1$, but not $f0$/$f1$, must be a black face (so to separate $of0$ and $of1$). However, this is a contradiction as $f2$ is adjacent to $f0$/$f1$ via a vertex and should have been included into the black component, which would violate the configuration of the two consecutive half-edges. 

For case (3) and (4), if $f0$ and $f1$ share an edge, then $of0$ and $of1$ must belong to the same black component. If $f0$ and $f1$ are the same, then $of0$ and $of1$ are of the same black component because they share a vertex. If $f0$ and $f1$ are different, then $of0$ and $of1$ are the same. $\square$
\end{proof}

\subsection{Proving that constraints (2), (3), and (4) of the IP problem have ensured that the original and the solved downsampled binary images have the same DAG}

We show a sketch of proof as follows.

\begin{proof}
Constraint (2) ensures that every nodes of the original DAG (i.e., every black and white components) exist in the downsampled DAG. 
Constraint (3) ensures that a black big-pixel is adjacent to either another black big-pixel of the same component, or to a white big-pixel. The same applies for white big-pixels. In other words, any two components of the same color won't touch each other.

However, the above two constraints still don't guarantee each original component remains one single component in the downsampled image. Now, constraint (4) dictates that every boundary between a black and a white component (which means all kinds of component-component boundaries as it is impossible to have two components of the same color to be adjacent to each other) remain a single close loop. This rules out the possibility of one original component become multiple connected components. $\square$
\end{proof}


\subsection{IP Problem Solving Outcomes Analysis}

In short, any solution that our method outputs will have the same topology as the input image. To be precise, there are 4 possible outcomes:

\begin{itemize}
  \item 1) and 2): The problem is solved optimally or sub-optimally (due to time limit) and a topologically correct solution with an optimal or sub-optimal objective value (in terms of pixel-wise similarity to the input image) is found.
  \item 3): The Gurobi solver finds the problem to be infeasible (w.r.t. the predefined ”coverage” dx / dy value - using a larger coverage value may, but not always, leads to more feasible solutions). Note that this is a certain conclusion (no feasible solution w.r.t. the coverage value would exist if the problem is found to be infeasible).
  \item 4): Gurobi could not find a solution nor being able to declare the problem to be infeasible within the time limit (we used 60 seconds). In our experiments, we found such cases to be rare (such cases are included in average time calculations throughout our tests).
\end{itemize}


\subsection{Ablation Study of the IP Formulation Constraints Design}
\label{sec:ablation}

\begin{figure}[t]
  \centering
  \includegraphics[width=0.9\linewidth]{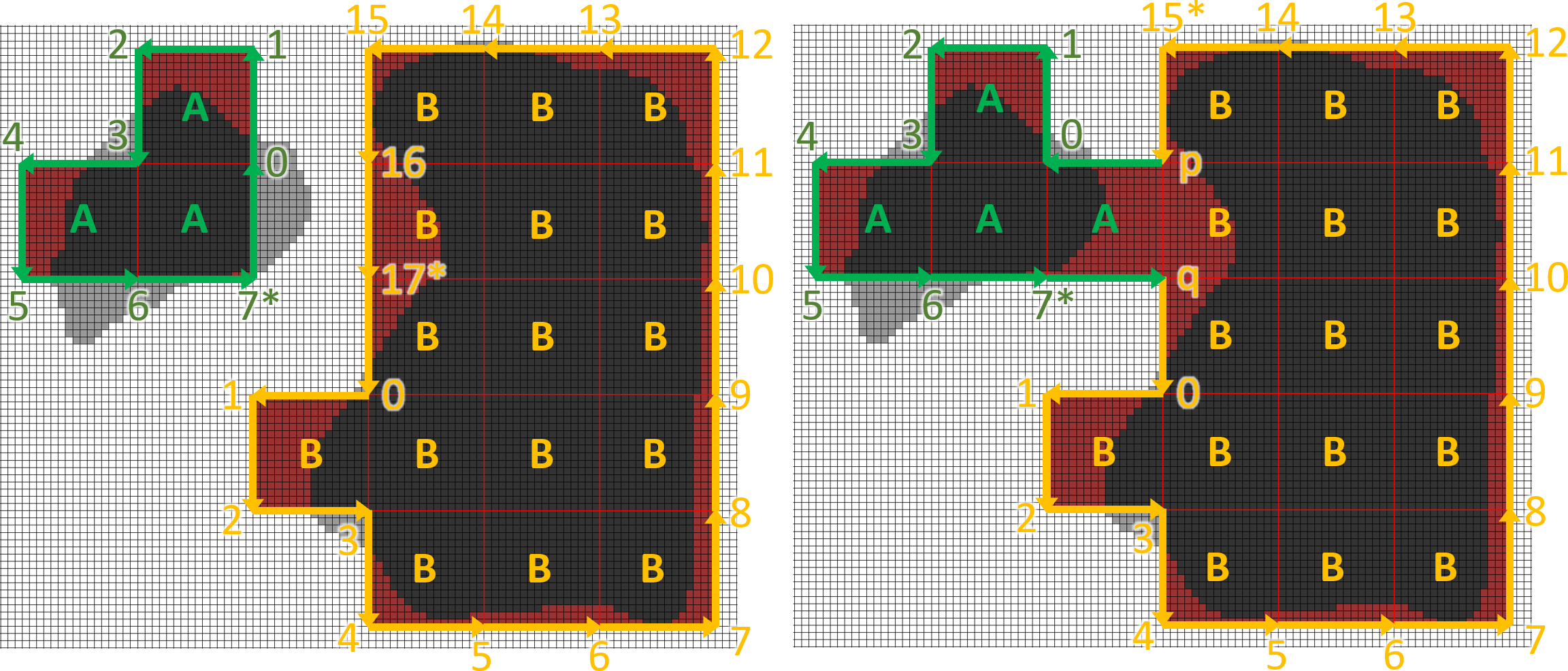}
  \caption{We show the necessity of constraint (3) in Section~\ref{sec:method}. In the right, we show that the two black components become a single one when constraint (3) is not enforced even when constraint (4) is being applied. The two components are represented by faces marked in A and B, respectively. Observe that in the right, constraint (4) actually still applies. This is because at point $p$ and $q$, none of the corner configurations are applicable, so their Boolean variables are not enumerated. This allowed the outer boundaries of component A and B to become non-closed intervals.}
  \label{fig:ablation}
\end{figure}

The necessities of constraint (1) (i.e., the image shall be fully covered by big-pixels without overlaps), (2) (i.e., every component shall appear), and (4) (the boundary between every pair of adjacent components shall be a single closed loop) in Section~\ref{sec:method} are apparent. However, it may not be clear if constraint (3) is needed when constraint (4) is already in place. We show a counter example and a discussion in Figure~\ref{fig:ablation}.


\subsection{More CNCB Dataset Downsampling Qualitative Results}
\label{sec:result_more_CNCB}

\begin{figure}[t]
	\begin{minipage}{1\linewidth}
		\vspace{2pt}
		\centerline{\includegraphics[width=\textwidth]{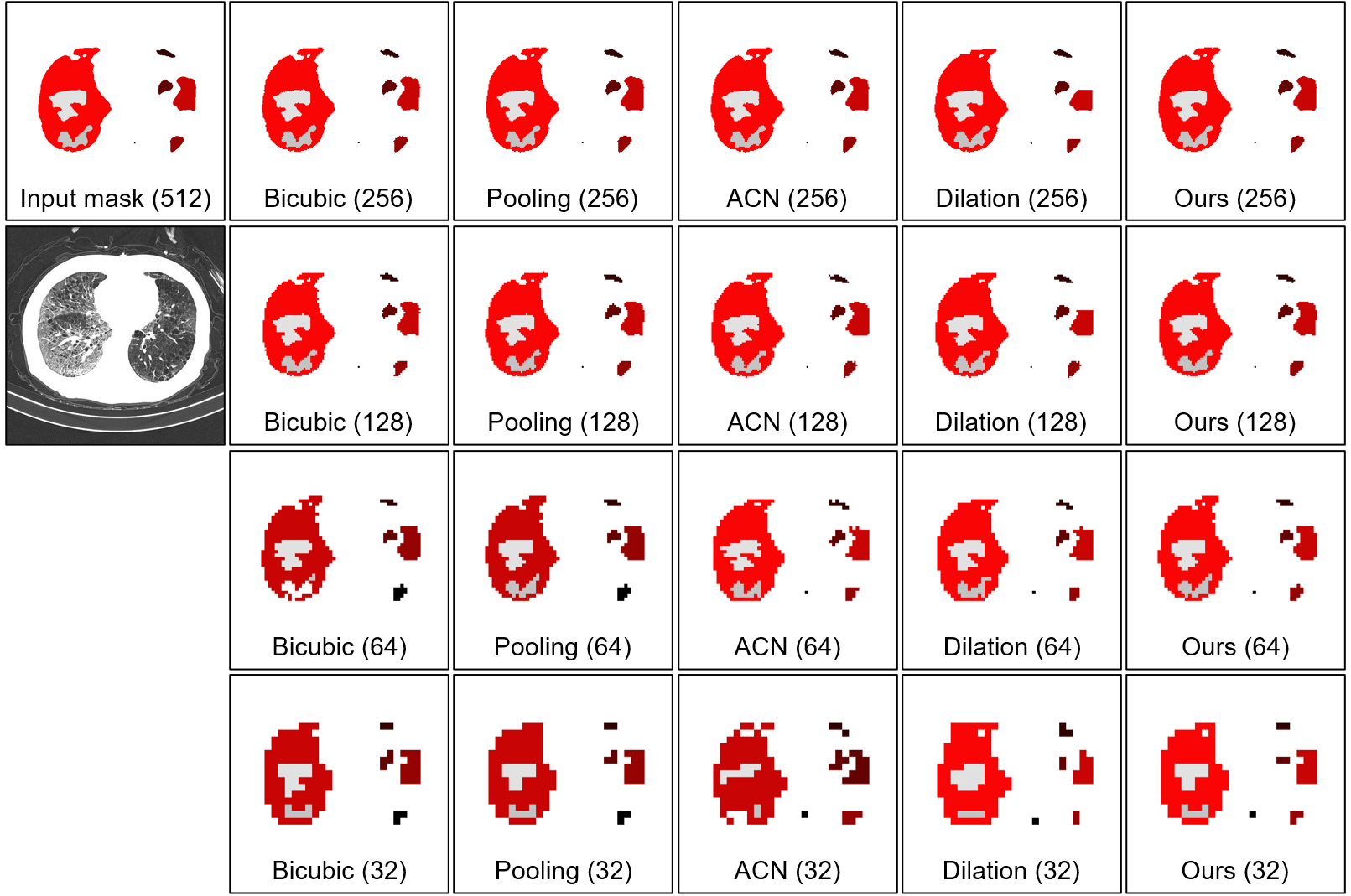}}
			\vspace{2pt}
		\centerline{\includegraphics[width=\textwidth]{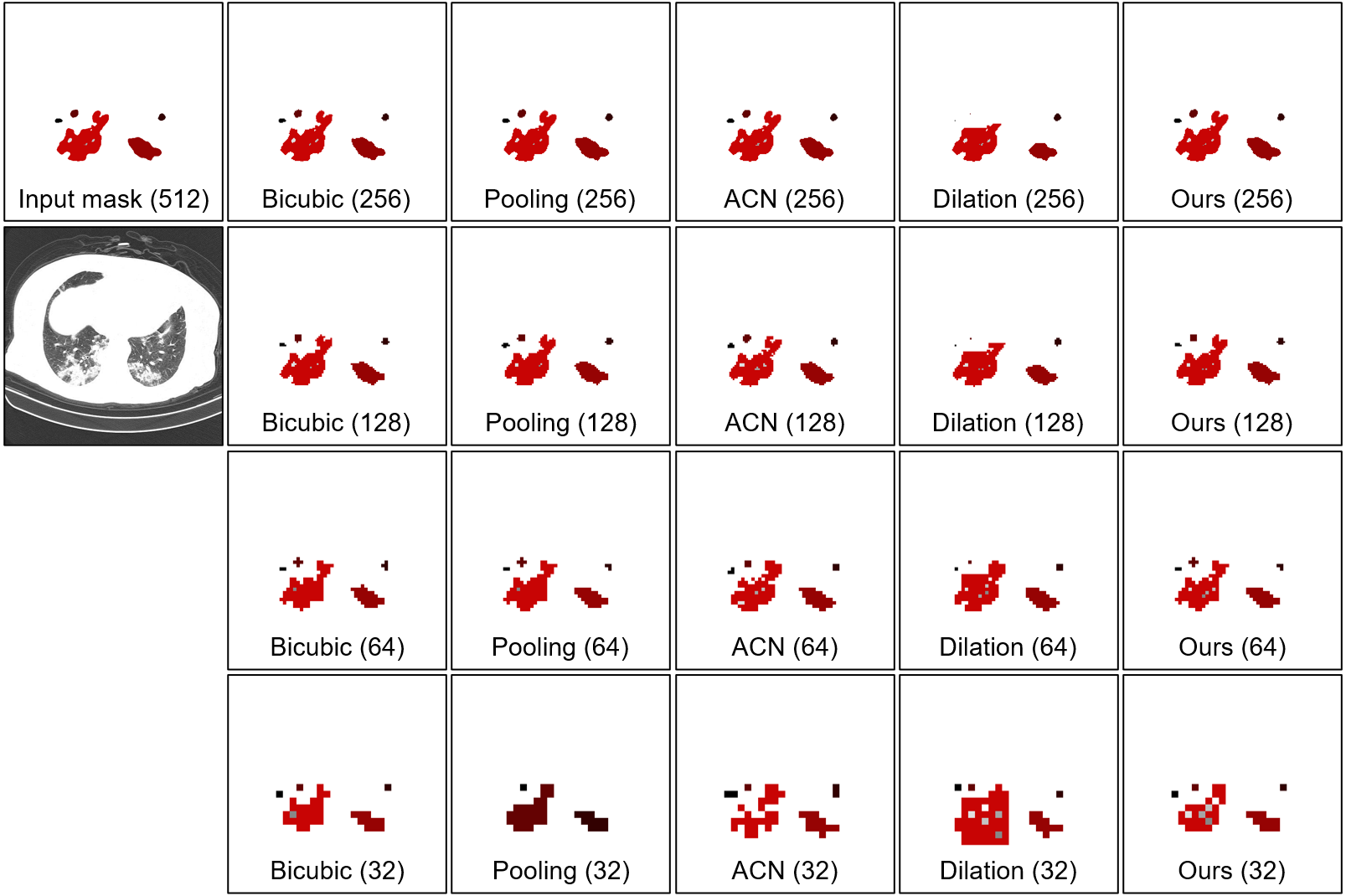}}
	\end{minipage}
	\caption{CNCB dataset downsampling results at all factors (2, 4, 8, and 16). Observe that results done by non topology-preserving methods often have altered connected components, especially at large factors (16).}
    \label{fig:medical_result2}
\end{figure}

\begin{figure}[t]
	\begin{minipage}{1\linewidth}
		\vspace{2pt}
		\centerline{\includegraphics[width=\textwidth]{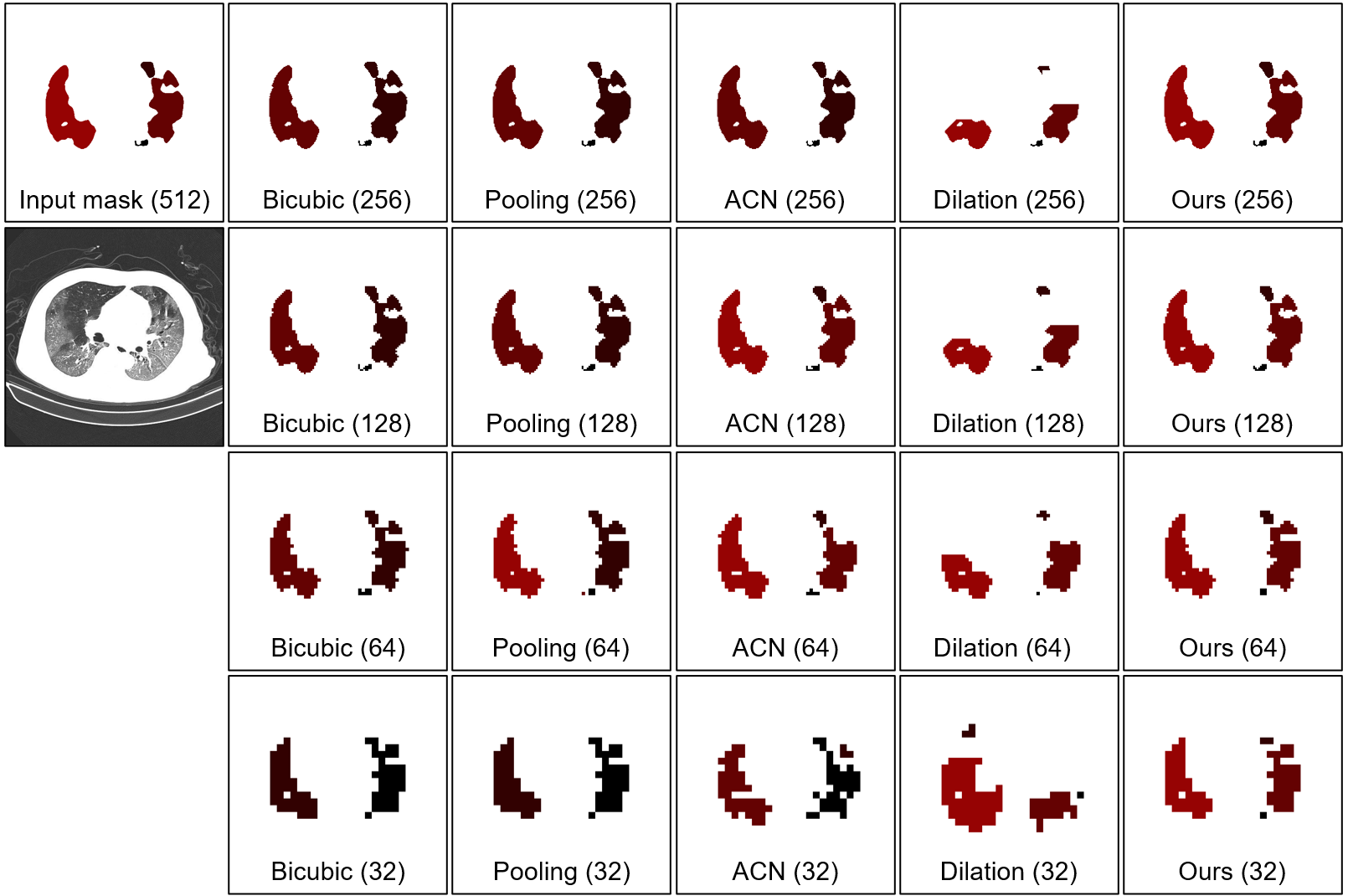}}
			\vspace{2pt}
		\centerline{\includegraphics[width=\textwidth]{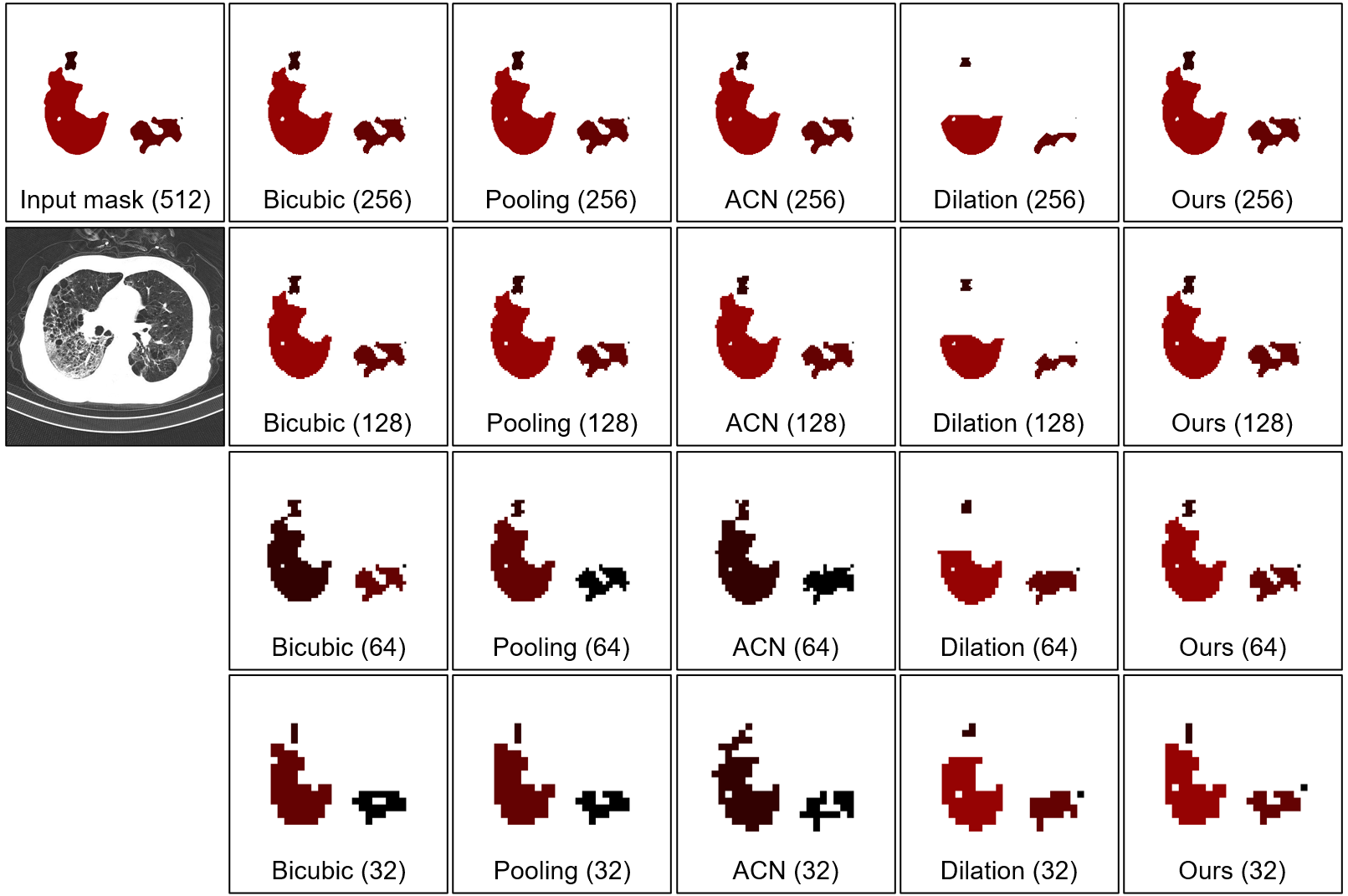}}
	\end{minipage}
	\caption{CNCB dataset downsampling results at all factors (2, 4, 8, and 16).}
    \label{fig:medical_result3}
\end{figure}

\begin{figure}[t]
	\begin{minipage}{1\linewidth}
		\vspace{2pt}
		\centerline{\includegraphics[width=\textwidth]{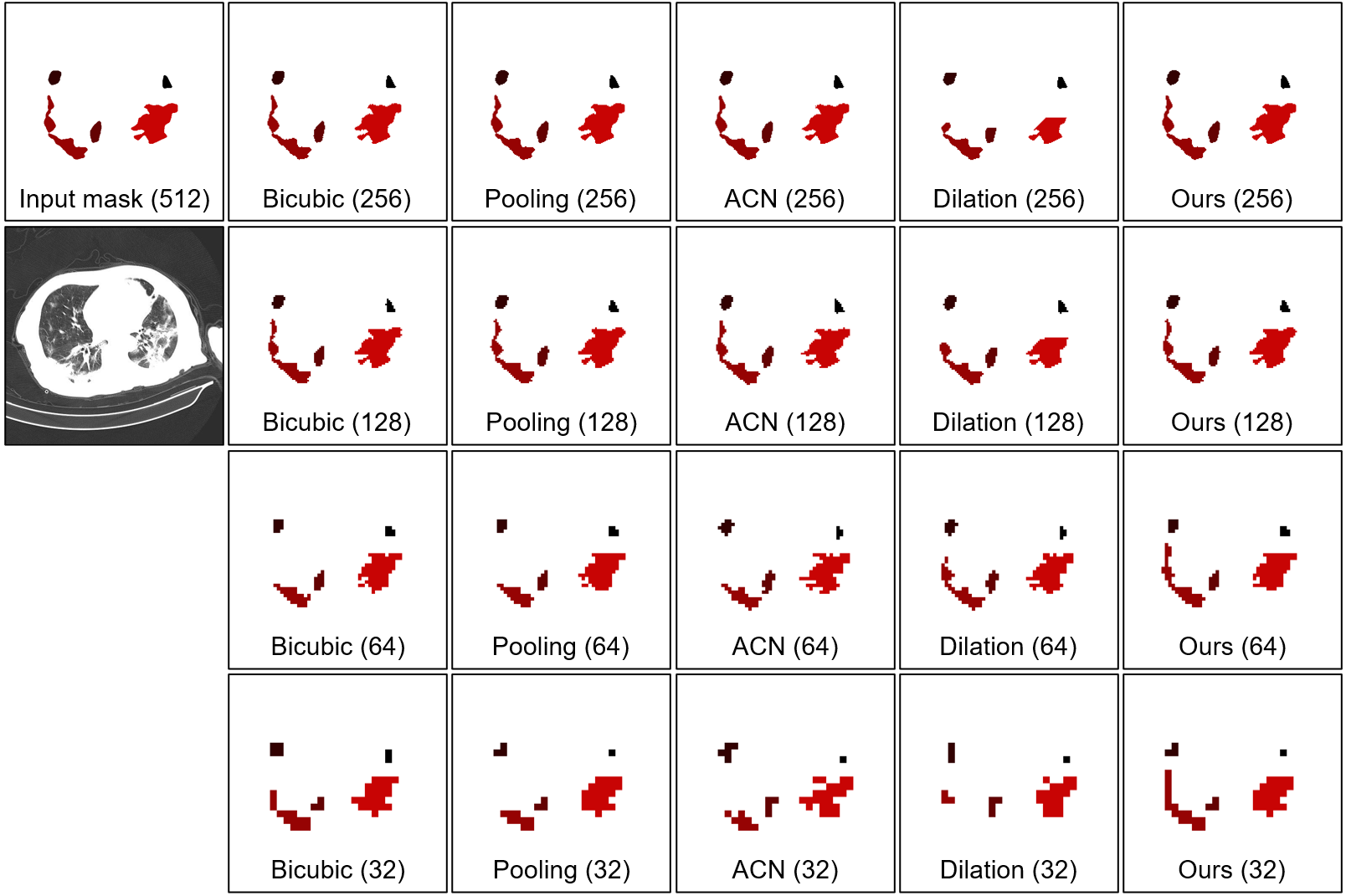}}
			\vspace{2pt}
		\centerline{\includegraphics[width=\textwidth]{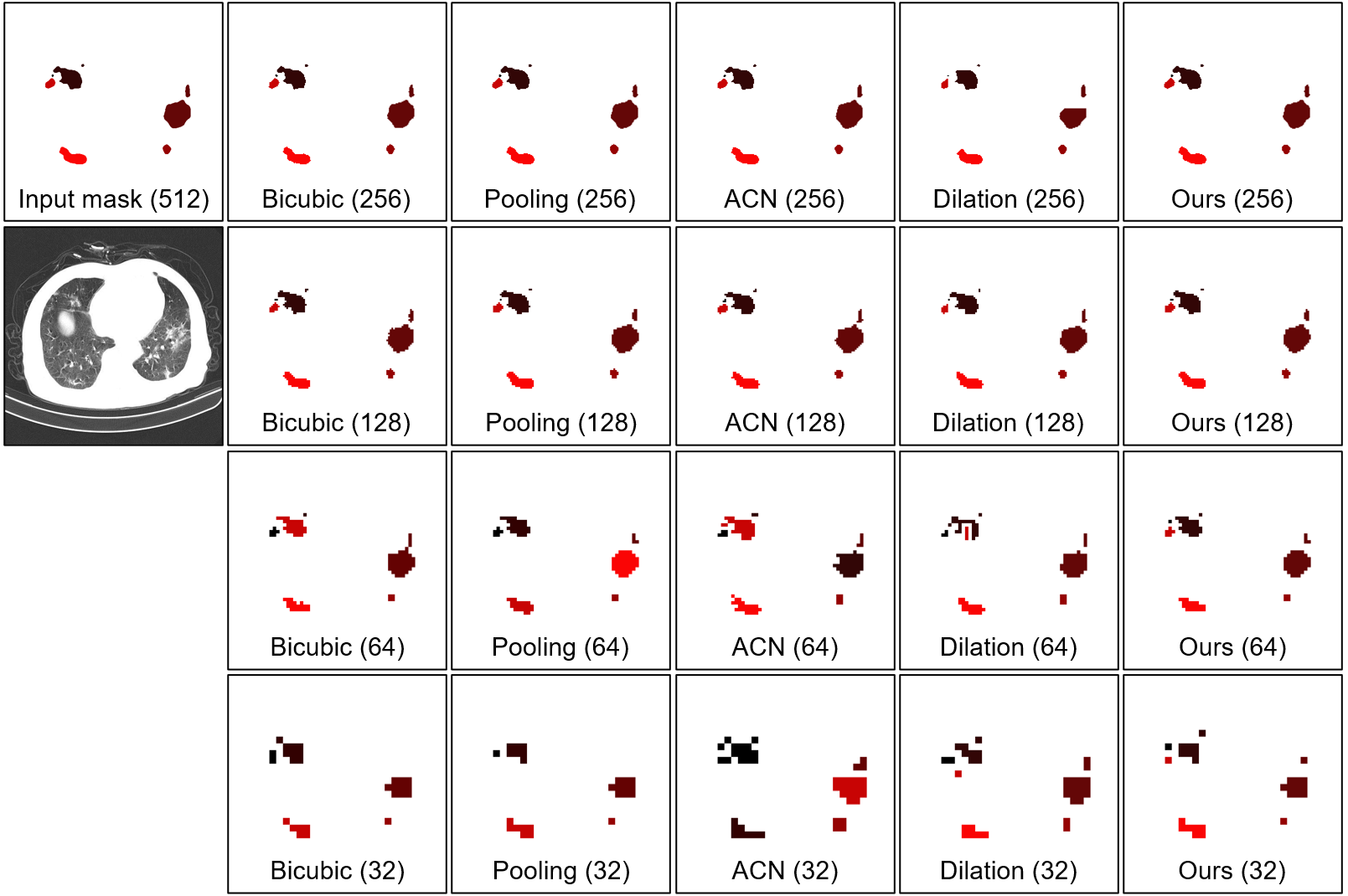}}
	\end{minipage}
	\caption{CNCB dataset downsampling results at all factors (2, 4, 8, and 16). }
    \label{fig:medical_result4}
\end{figure}

We show more qualitative results on the CNCB dataset with more downsampling factors (from 2 to 16) in Figure~\ref{fig:medical_result2}, Figure~\ref{fig:medical_result3}, and Figure~\ref{fig:medical_result4}. We point out that at big downsampling factors, results done by non topology-preserving methods (e.g., bicubic, pooling, and ACN) often have altered connected components, such as small components (or holes) got erased, multiple components/holes got connected, or one component/hole got separated into many.

\subsection{Persistent Diagram Comparisons}
\label{sec:result_persistent_diagram}

\begin{figure}[]
  \centering
  \includegraphics[width=1\linewidth]{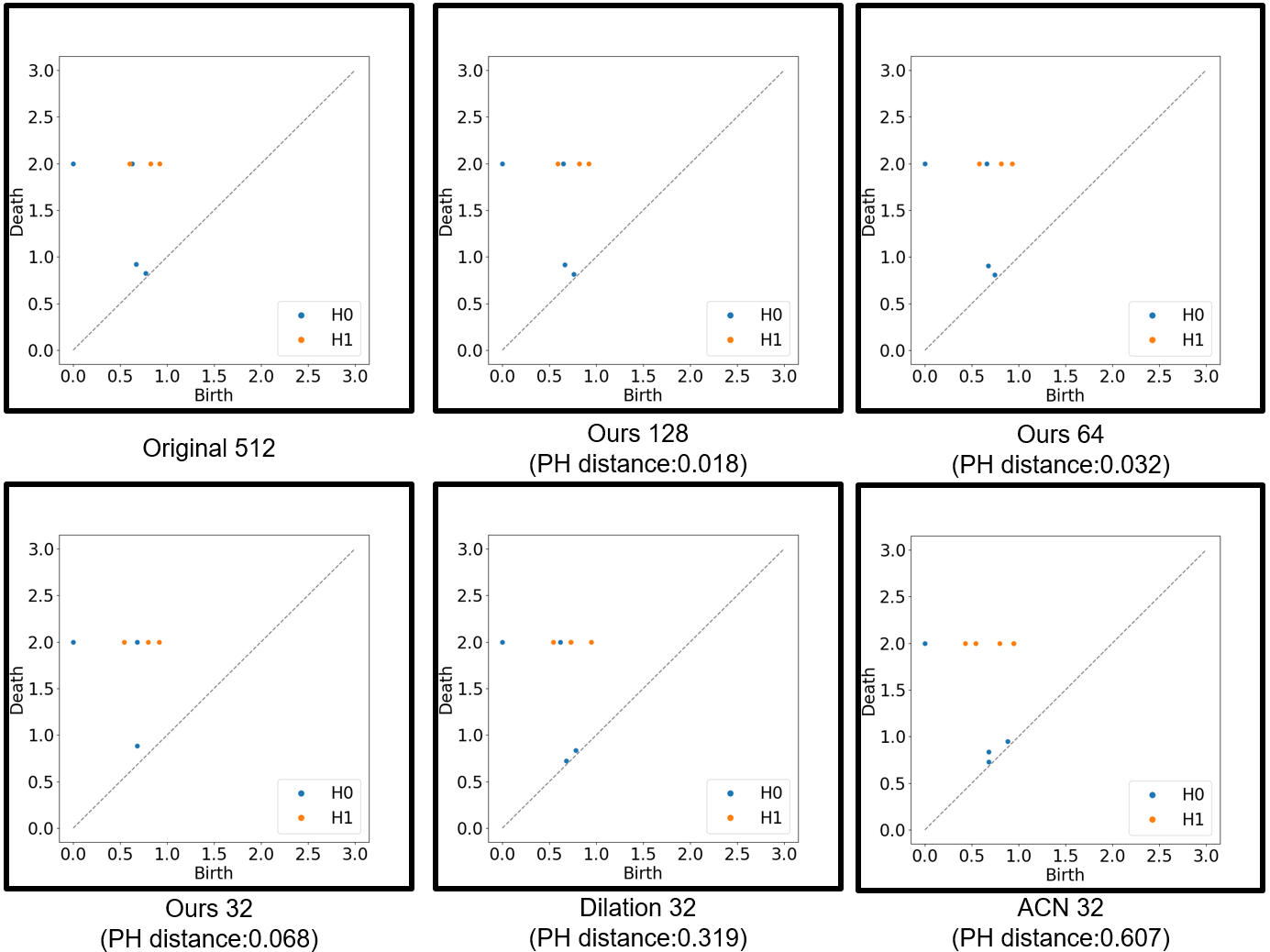}
  \caption{Persistent diagrams of an original 512x512 binary image and its downsampled versions by our method (to 128x128, 64x64, and 32x32), and some results done by other methods (32x32).}
  \label{fig:persistent_diagram}
\end{figure}

In Section~\ref{sec:result}, we have shown quantitative comparisons (in terms of PH distances) of persistent diagrams of original binary images and downsampled images by our method and other methods. To have an qualitative comparison, in Figure~\ref{fig:persistent_diagram}, we show persistent diagrams of an original 512x512 binary image and its downsampled versions by our method, and some results done by other methods. We see that persistent diagrams of our downsampled images are visually similar to the original. In comparison, persistent diagrams of results done by other methods, having significantly higher PH distances, look more different to the original.

\subsection{Shortest Path Comparisons}
\label{sec:result_shortest_path}

\begin{figure}[]
  \centering
  \includegraphics[width=1\linewidth]{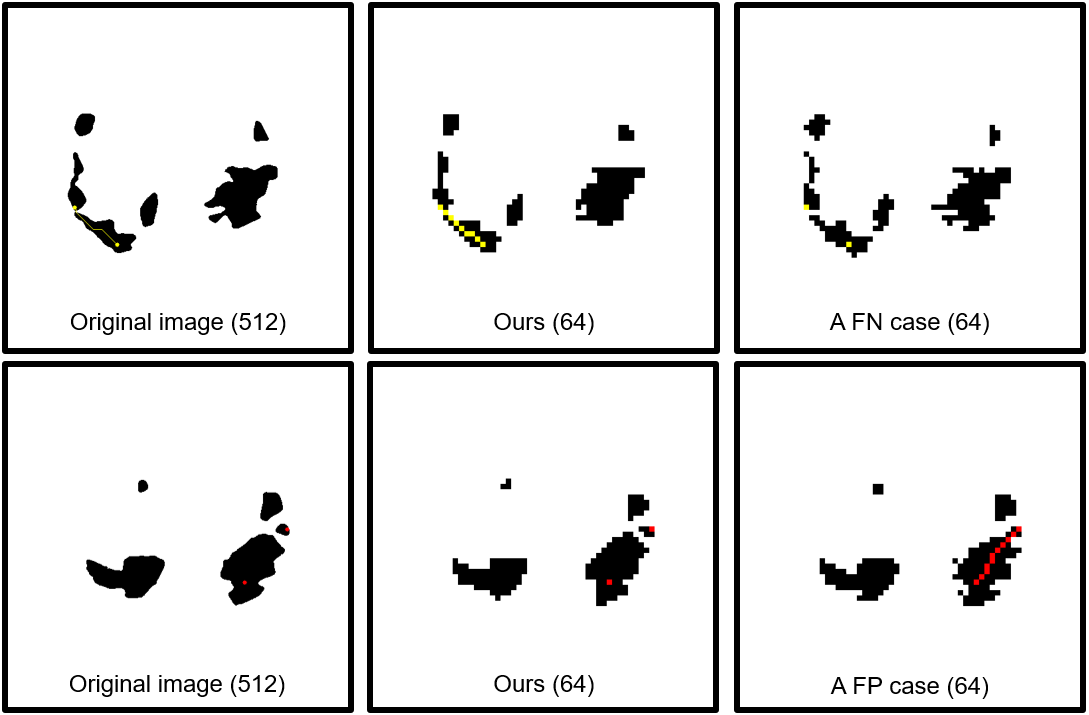}
  \caption{Left: a shortest path (between two yellow pixels) and a pair of pixels without a path (red) in an original image. Middle: shortest path finding results in our downsampled image. Right: we show a false negative (FN) case because the component containing the two yellow pixels now becomes two. We also show a false positive (FP) case because the two red pixels now belong to the same component and a path emerges.}
  \label{fig:shortest_path}
\end{figure}

In Figure~\ref{fig:shortest_path}, we show approximate shortest paths in our downsampled images. We also show a false negative (FN) case and a false positive (FP) case, both led to highly incorrect shortest path estimations, that may happen in topologically incorrect downsampled images.

\subsection{Bigger Resolution Inputs Testing Results}
\label{sec:result_bigger}

\begin{figure}[t]
	\begin{minipage}{1\linewidth}
		\vspace{2pt}
		\centerline{\includegraphics[width=\textwidth]{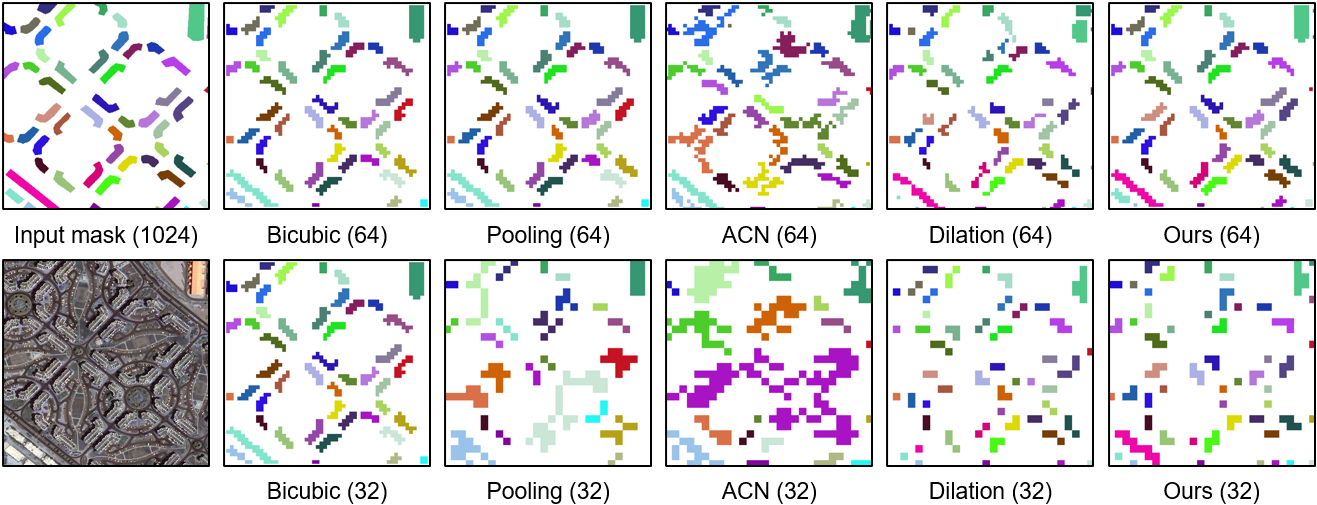}}
			\vspace{2pt}
		\centerline{\includegraphics[width=\textwidth]{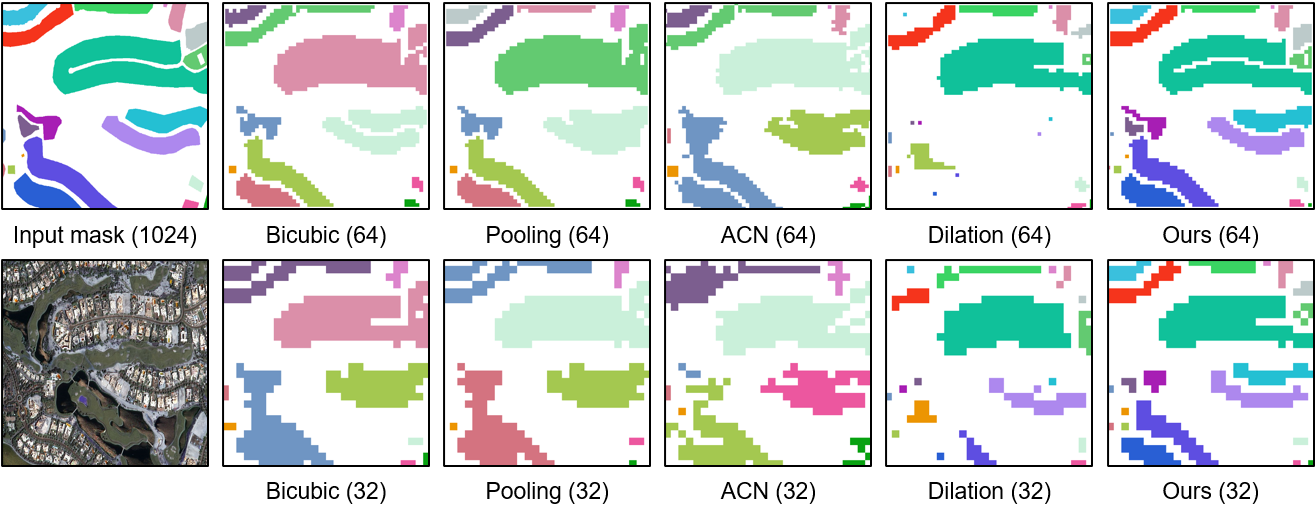}}
	\end{minipage}
	\caption{Downsampling 1024x1024 images to 64x64 (factor 16) and 32x32 (factor 32).}
    \label{fig:big_aerial_result}
\end{figure}

To see how our method performs on bigger ($\ge$ 512x512) input binary images, we tested on two big images found in the aerial image dataset. Statistics are shown in Table~\ref{tab:big_aerial_result} and downsampling results are shown in Figure~\ref{fig:big_aerial_result}. By comparing to Table~\ref{tab:aerial_result}, we see that the computational cost of downsampling a 1024x1024 image to 64x64 is only slightly higher then converting a 512x512 image to 64x64 (from 3.494 sec to 3.647 sec). The same can be said about comparing converting 1024x1024 and 512x125 images to 32x32 images (from 5.348 sec to 5.550 sec).

\begin{table}
\scriptsize
\begin{center}
\caption{\label{tab:big_aerial_result}Quantitative and speed comparisons of different downsampling methods on two bigger (1024x1024) input binary images in the aerial image dataset to different sizes. {\color{red}Best} and {\color{blue}second-best} results are marked in red and blue, respectively.}
\label{table:big_aerial}
\resizebox{\textwidth}{!}{
\begin{tabular}{l|ccccc|ccccc}

 & \multicolumn{5}{c|}{1024x1024 to 64x64 (factor 16)} & \multicolumn{5}{c}{1024x1024 to 32x32 (factor 32)} \\

\hline
Method & $\uparrow$ IoU & $\uparrow$ Dice & \makecell{$\downarrow$ Betti \\ num. error} &  \makecell{$\downarrow$ PH \\ distance} & \makecell{$\downarrow$ Avg. \\ time (s)} & $\uparrow$ IoU & $\uparrow$ Dice & \makecell{$\downarrow$ Betti \\ num. error} &  \makecell{$\downarrow$ PH \\ distance} & \makecell{$\downarrow$ Avg. \\ time (s)}\\
\hline
Bicubic & \textcolor{blue}{76.62\%} & \textcolor{blue}{86.56\%} & 5.00
& 0.158 & \textcolor{red}{0.004} & \textcolor{red}{59.87\%} & \textcolor{red}{74.33\%} & \textcolor{blue}{10.58} & 0.179 & \textcolor{red}{0.004}\\
Pooling & \textcolor{red}{77.00\%} & \textcolor{red}{86.81\%} & \textcolor{blue}{4.42} & 0.175 & \textcolor{blue}{0.066} & \textcolor{blue}{59.53\%} & \textcolor{blue}{73.83\%} & 10.67 & 0.202 & \textcolor{blue}{0.038}\\
ACN~\cite{decenciere2007adaptive}  & 62.39\% & 76.42\% & 9.17 & 0.212 & 0.801 & 43.83\% & 60.49\& & 12.17 & 0.229 & 0.879\\
Dilation & 56.56\% & 71.50\% & \textcolor{red}{0.00} & \textcolor{blue}{0.107} & 2.232 & 37.12\% & 53.31\% & \hphantom{1}\textcolor{red}{0.00} & \textcolor{blue}{0.159} & 0.674\\
Ours & 76.52\% & 86.51\% & \textcolor{red}{0.00} & \textcolor{red}{0.075} & 3.647& 54.43\& & 69.69\% & \hphantom{1}\textcolor{red}{0.00} & \textcolor{red}{0.095} & 5.550\\
\hline
\end{tabular}
}
\end{center}
\end{table}

\subsection{Anisotropic Downsampling Results}
\label{sec:result_nonsquare}

\begin{figure}[]
  \centering
  \includegraphics[width=1\linewidth]{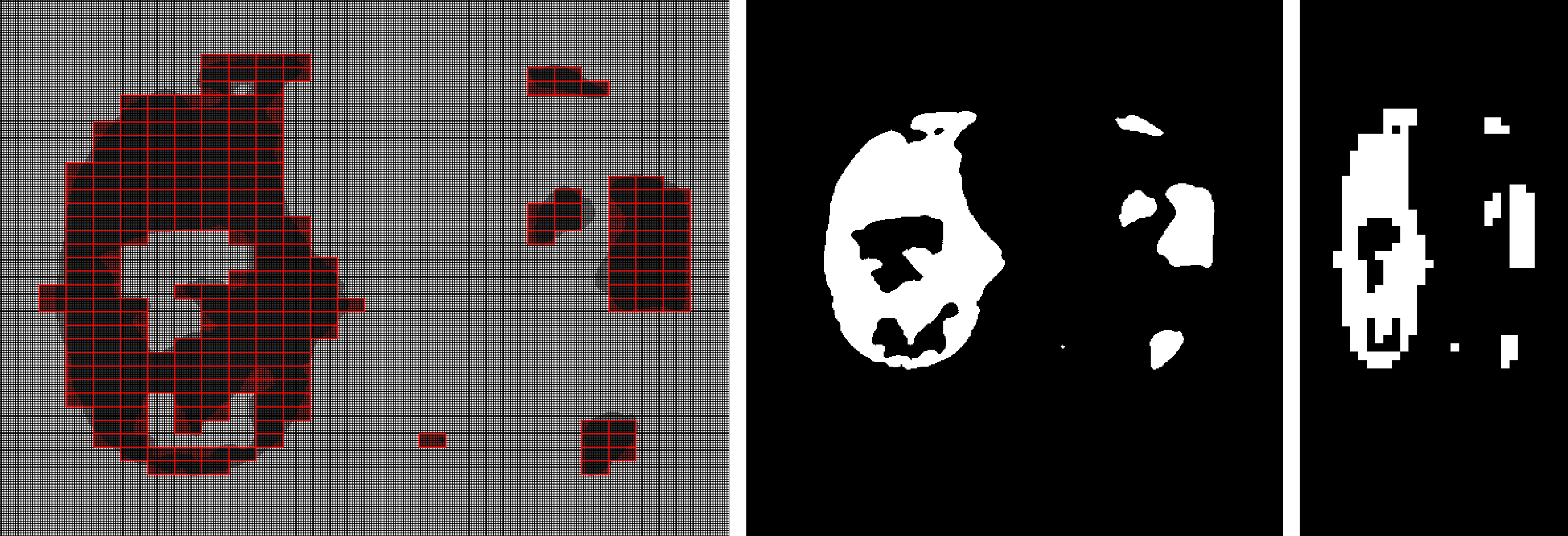}
  \caption{Left: visualizing a downsampling result with factor 16 in width and 8 in height. Observe that each big-pixel is a 16-by-8 rectangle. Middle: original image, Right: downsampled image.}
  \label{fig:nonsquare}
\end{figure}

Our method supports anisotropic downsampling by using different downsampling factors in width and height. In Figure~\ref{fig:nonsquare}, we show one such result of downsampling a 512x512 image into a 32x64 one with downsampling factor 16 in width and 8 in height.

\end{document}